\documentclass[nohyperref]{article}

\usepackage{microtype}
\usepackage{graphicx}
\usepackage{subfigure}
\usepackage{booktabs} 

\usepackage{hyperref}

\usepackage[accepted]{icml2023}

\usepackage{amsmath}
\usepackage{amssymb}
\usepackage{mathtools}
\usepackage{amsthm}

\usepackage[capitalize,noabbrev]{cleveref}

\theoremstyle{plain}
\newtheorem{theorem}{Theorem}[section]

\theoremstyle{definition}

\theoremstyle{remark}
\newtheorem{remark}[theorem]{Remark}

\usepackage[textsize=tiny]{todonotes}

\icmltitlerunning{\textsc{Moccasin}: Efficient Tensor Rematerialization for Neural Networks}

\usepackage{multirow}

\begin{document}

\twocolumn[
\icmltitle{\textsc{Moccasin}: Efficient Tensor Rematerialization for Neural Networks}




\begin{icmlauthorlist}
\icmlauthor{Burak Bartan}{qualcomm}
\icmlauthor{Haoming Li}{usc}
\icmlauthor{Harris Teague}{qualcomm}
\icmlauthor{Christopher Lott}{qualcomm}
\icmlauthor{Bistra Dilkina}{usc}
\end{icmlauthorlist}

\icmlaffiliation{qualcomm}{Qualcomm AI Research, Qualcomm AI Research is an initiative of Qualcomm Technologies, Inc.}
\icmlaffiliation{usc}{University of Southern California, Los Angeles, USA}

\icmlcorrespondingauthor{Burak Bartan}{bbartan AT qti.qualcomm.com}

\icmlkeywords{Machine learning, rematerialization, integer programming}

\vskip 0.3in
]



\printAffiliationsAndNotice{}  

\begin{abstract}
The deployment and training of neural networks on edge computing devices pose many challenges. The low memory nature of edge devices is often one of the biggest limiting factors encountered in the deployment of large neural network models. Tensor rematerialization or recompute is a way to address high memory requirements for neural network training and inference. In this paper we consider the problem of execution time minimization of compute graphs subject to a memory budget. In particular, we develop a new constraint programming formulation called \textsc{Moccasin} with only $O(n)$ integer variables, where $n$ is the number of nodes in the compute graph. This is a significant improvement over the works in the recent literature that propose formulations with $O(n^2)$ Boolean variables. We present numerical studies that show that our approach is up to an order of magnitude faster than recent work especially for large-scale graphs.
\end{abstract}
\section{Introduction} \label{sec:introduction}

The need for efficient deep neural network (DNN) computations for both training and inference continues to grow.  Most compute architectures for DNNs make use of a limited amount of fast, \textit{local memory} in conjunction with a much larger amount of \textit{global memory} that is significantly slower to access.  For example, accelerators for mobile devices today will dedicate several MB of local memory (per core) for storage of intermediate output tensors before resorting to the devices' main memory; training of large DNNs are done primarily on GPU's local GDDR memory before offloading tensors to DDR memory on the motherboard. However, there are many DNNs in practice for which the local memory is not enough to store all the required intermediate outputs.  Effective optimization of the local memory footprint during the computation can make a difference in meeting latency targets for computations since global memory accesses can be too slow (\citealp[Sec. V]{sze2017efficient}). In addition, the trend toward on-device training will further stress the limited resources due to the need to store intermediate data for gradient back-propagation \cite{tinyML_OnDevice_Form}.

\begin{figure}
    \centering
    \includegraphics[width=0.65\columnwidth]{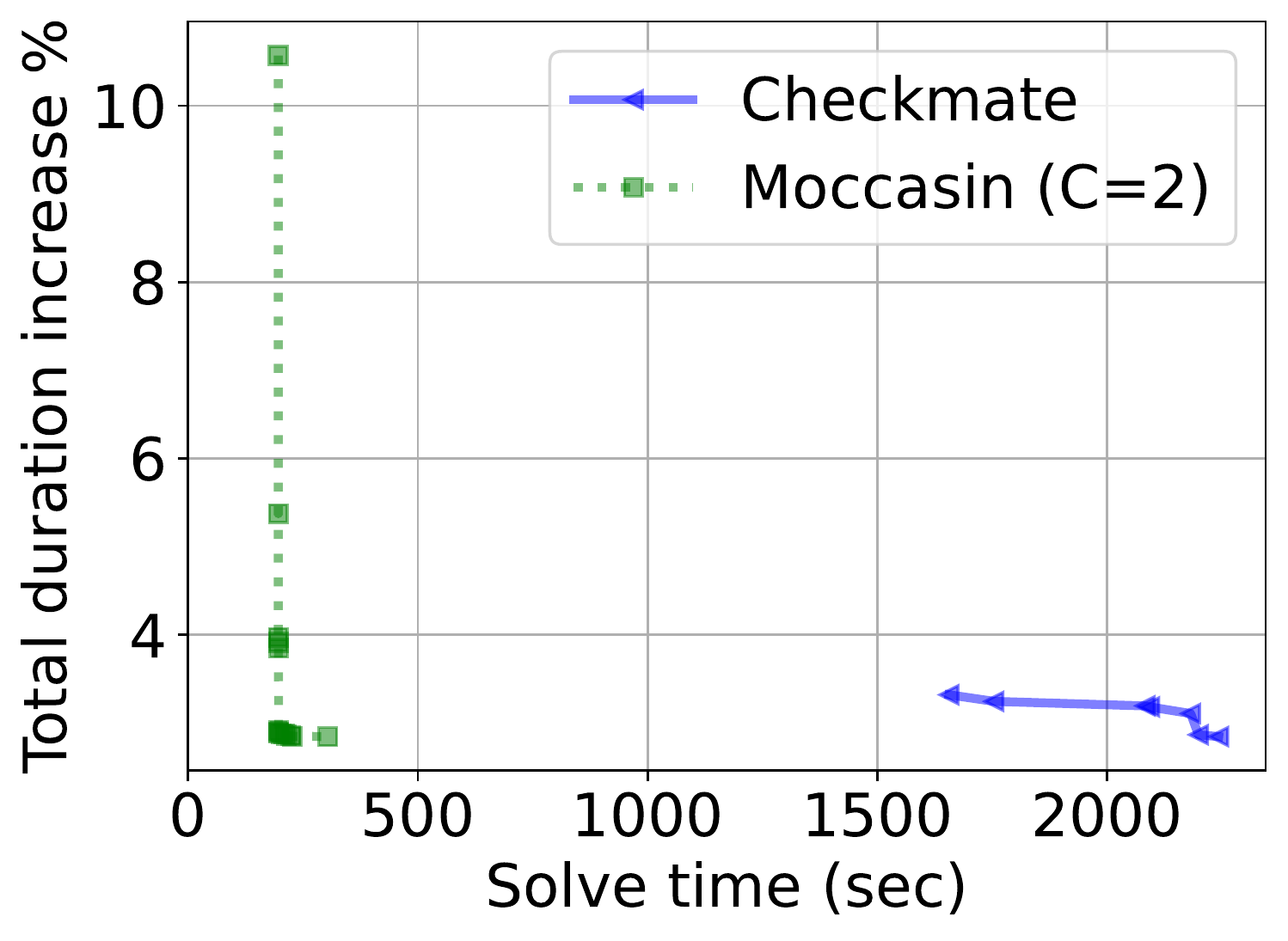}
    \vspace{-2mm}
    \caption{Percentage of total duration increase against solve time in seconds for our proposed method (\textsc{Moccasin}) and \textsc{Checkmate} \cite{jain2020checkmate} on a real-world graph with $n=442$ nodes and $m=1247$ edges. The memory budget is set to $80\%$ of the peak memory for the initial topological order without rematerialization. The time until the first data point in each curve is spent on presolve.}
    \label{fig:comparison_intro}
\end{figure}

There are several techniques for DNN computation sequencing that can be used to manage the local memory footprint, and a well-developed toolchain for this purpose is likely to employ multiple methods working together. For example, \textit{operation sequencing} is a technique studied in \cite{topoformer} and its references.  Our paper focuses on another technique 
called \textit{rematerialization}.  When input data is needed for a compute operation, it can be read from local memory, or alternatively be recomputed (rematerialized). In some cases the extra compute can prove to be advantageous relative to occupying valuable local memory with the data until it is ready to be used, with the latter leaving a larger memory footprint.  Deciding what operations to rematerialize (or not) in order to optimize latency and memory-use is a PSPACE-complete problem \cite{gilbert1979pebbling}.  Further, we often need a method that can tackle large computation graphs with acceptable “compile time”, thus need an approach that scales well with graph size.

While there is a history of rematerialization research for managing register use by compilers \cite{kubota1998fortran77,briggs1992rematerialization}, more recently, these techniques have been further developed and studied for use in DNNs, where the size of computation graphs and the amount of data movement can be extremely large \cite{jain2020checkmate, patil2022poet, schuler2022xengine}. These works formulate a mixed integer linear program (MILP) with a Boolean variable for each use of a compute node’s output – either read from memory or rematerialized.  We have found that this formulation has limitations when attempting to scale to large graphs due to the need for $O(n^2)$ Boolean variables (where $n$ is the number of nodes in the computation graph).  While linear relaxation followed by rounding is an approximation that is leveraged in these works, we have found that these rounded solutions can be far from optimal, limiting the applicability of this approach.

The general rematerialization problem is stated here, and will be made more explicit in Section \ref{sec:formulation}.  Given a directed acyclic graph (DAG), $G=(V, E)$, with $|V|=n$ and $|E|=m$, let nodes $v \in V$ represent compute operations and directed edges $(u,v) \in E$ represent data dependencies such that the output tensors of all the nodes $\{w: (w,v)\in E \}$ (predecessors of $v$) are required to be present in local memory before computing $v$.  Let $seq(G)$ be a \textit{rematerialization sequence} of nodes (ordered list) that contains each node at least once. The rematerialization problem statement is then as follows:

\textsc{Memory-constrained Computation Graph \mbox{Sequencing} with Rematerialization}
\begin{align*}
    &\underset{seq(G)}{\mbox{minimize}} \,\quad  \text{total execution duration} \\
    &\mbox{subject to} \quad  \\ 
    \quad & \quad - seq(G) \text{ meets the data dependencies of } G\\
    \quad & \quad - \text{peak memory footprint of $seq(G)$} \\
    \quad & \quad\quad\quad \text{$\leq$ local memory capacity $M$} 
\end{align*}
This definition is intentionally high-level and lacking some details that will be provided later. 
Since our later formulations do not directly optimize the sequence but rather optimize other variables from which memory footprints and the final sequence are computed, we skip further description of such details in this section.

Our work introduces a new optimization model formulation that defines the problem variables as \textit{retention intervals}, allowing use of only $O(n)$ integer variables.  We then leverage constrained programming (CP), allowing complex data dependency and cumulative memory threshold constraints to be enforced easily and efficiently during optimization.  We further show that our proposed approach demonstrates significant scaling improvements relative to the MILP formulation of \textsc{Checkmate} \cite{jain2020checkmate}. Figure \ref{fig:comparison_intro} provides a numerical comparison of the two methods. 

\subsection{Related Work} 
The problem of rematerialization has its roots in compiler optimization \cite{colombet2015studying,lozano2019survey,lozano2019combinatorial} and automatic differentiation with checkpointing \cite{kubota1998fortran77,griewank2008evaluating,griewank2000algorithm}. Recently, there is a stream of works on rematerialization in machine learning, motivated by the need to train large DNN on GPU with limited memory capacity \cite{kumar2019efficient,kusumoto2019graph,beaumont2021efficient,chen2016training,kirisame2021dynamic,mostafa2022sequential,huang2019gpipe}; most relevant to us are the works based on combinatorial optimization \cite{jain2020checkmate,patil2022poet,schuler2022xengine}.

\label{sec:related_work}
The scope of our work is most closely aligned with that of \cite{jain2020checkmate}, where they develop a mixed integer linear program (MILP) for duration minimization under memory constraints referred to as \textsc{Checkmate}.  The formulation of \textsc{Checkmate} defines Boolean matrix variables $R,S,F$ for node/tensor recomputation, local memory storage, and deallocation, respectively, resulting in $O(n^2)$ Boolean variables. While the paper demonstrates the capability of rematerialization to meet memory limits while minimizing duration, the paper does not sufficiently address the problem complexity and scaling to larger graphs.  The authors do mention this issue and offer a method of linear relaxation of the Boolean variables followed by 2-stage rounding to accelerate finding a solution.  However, our experiments show that for many graphs, this rounding approach produces results that are far from optimal, and often the rounded solution does not meet the memory threshold and is thus infeasible (see Section \ref{sec:numerical_results}).

As a consequence of formulating the problem using matrices of variables where the columns represent specific nodes (permutations are not supported), in \textsc{Checkmate} all solutions are constrained to follow an input sequence of the graph nodes without rematerialization.  We call this the ``input topological order".  This input must be a valid topological order for solutions that meet the node precedence requirements to be possible.  The question then arises, ``what input topological order should be used?"  Indeed this constraint significantly reduces the space of potential valid rematerialization sequences so can be valuable to limit algorithm complexity.  However, as shown in \cite{topoformer}, there can be a wide variability of peak memory footprint for different sequences (without rematerialization).  

One of the attractive features of our new formulation is that it does not require an input topological order.  However, when studying the graphs used in \cite{jain2020checkmate}, we found no variations in the peak memory footprint across 50 randomly generated topological orders for each graph (without rematerialization).  To compare directly with prior methods, we add the input topological order constraint (see Section \ref{sec:enforce_topo}) and use this for all of our experiments. Removing the input topological order constraint and/or studying the impact of different input topological orders on solution quality and solve time are an interesting topic for future research.

A follow-on to \textsc{Checkmate} \cite{patil2022poet} recognizes that memory optimization may combine rematerialization with paging -- strategically reading and writing intermediate data to external memory.  While these external memory accesses are slow, they still may be required if the available local memory is too small to fit any rematerialization solution of practical total duration.  This work, however, still suffers from a scaling issue as it inherits the $O(n^2)$ Boolean variable formulation.  An extension of our formulation to incorporate paging would be valuable future research. Another extension is \cite{schuler2022xengine} which builds on the \textsc{Checkmate} formulation to allow making rematerialization decisions across two heterogeneous cores. This extension also suffers from the same fundamental scaling issue.

In \cite{kumar2019efficient}, a novel method is introduced for finding a valid rematerialization sequence (one that meets the required memory threshold) by performing divide-and-conquer using tree decomposition.  This work focuses on proving an attractive complexity for generating such a sequence.  However, it is of limited use in practice since it does not contain any explicit mechanism for minimizing total duration of the final sequence produced by the algorithm.

The potential for reducing peak local memory footprint using rematerialization depends on the computation graph topology.  For example, a line graph (string of nodes connected by single edges) offers no potential for improvement since the local memory footprint at each node does not depend on past deallocations.  In contrast, a simple U-net typically allows significant opportunities for footprint savings.  In general, we observe that networks with ``long skip connections" are the topologies that exhibit more potential for rematerialization gains.  DNN training computation graphs have a ``U-net-like" structure since they have forward and backward paths with edges that cross between the two to support gradient computation in the backward path.  \cite{jain2020checkmate} only experiments with training graphs.  In contrast, we study rematerialization on both training and inference graphs, and find compelling rematerialization gains also for inference graphs, particularly ones with complex interconnect topology.

\subsection{Contributions}
\begin{itemize}
    \item \emph{Retention intervals formulation of rematerialization:} We introduce \textsc{Moccasin}, a new formulation with significantly better scaling of solution complexity to large graphs.  We demonstrate that this formulation can be solved effectively using \textit{constraint programming} (CP).
    \item \emph{Constrained number of allowed rematerializations:}  We introduce parameter $C_v$ that defines the maximum number of times a node $v$ can be computed in the final sequence.  We demonstrate empirically that this complexity reduction retains solution quality even for very small values of $C_v$.
    \item \emph{Compare and contrast approaches:}  We provide comparison of solution speed for \textsc{Checkmate} vs \textsc{Moccasin} and demonstrate equivalence of solutions.
    \item \emph{Impact of memory limit:}  Prior works make simple assumptions on local memory limit in their evaluation.  We show the impact of a range of local memory limits on solution speed and final solution value. 
    
\end{itemize}

\section{Retention Intervals Formulation} \label{sec:formulation}

The main building block of our formulation is the concept of \textit{output retention intervals} which, as we will show, simplifies the problem formulation greatly. For each node in the computation graph, we will define intervals that indicate the retention of its output in local memory. More precisely, each rematerialization of a node will have its own interval. Here, with a misuse of terminology, we use the term rematerialization for the first time that a node is computed as well. We will assume a node $v \in V$ is allowed to be rematerialized up to $C_v$ times, hence $C_v$ intervals will be defined for node $v$. The parameter $C_v$ could be considered a hyperparameter. We will show via numerical experiments that picking $C_v$ to be as small as $2$ for all of the nodes is often good enough in practice and is without loss of optimality (see Section \ref{sec:numerical_results}).
%
%



We define the retention intervals for each node $v\in V$ by its ``start" and ``end" times 
\begin{align*}
    s_v^i, e_v^i  \in \mathcal{D}, \,\, \forall i \in \{1,\dots,C_v\} \,,
\end{align*}
where the domain $\mathcal{D}$ is a set of integers of size $O(n)$ that will be defined explicitly in the subsequent subsections. A memory block of size $m_v$ is allocated at the start and deallocated at the end of the interval. 

\begin{figure}
\centering
\begin{tikzpicture}[node distance={15mm}, thick, main/.style = {draw, circle}] 
\node[main] (1) {$1$}; 
\node[main] (2) [right of=1] {$2$};
\node[main] (3) [right of=2] {$3$};
\node[main] (4) [right of=3] {$4$}; 
\draw[->] (1) -- (2); 
\draw[->] (2) -- (3); 
\draw[->] (3) -- (4); 
\draw[->] (1) to [out=30,in=150] (4); 
\end{tikzpicture}
\caption{Example graph with 4 nodes.}
\label{fig:4nodegraph}
\end{figure}
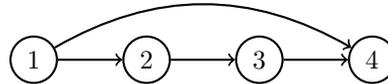

\begin{figure}
    \centering
    \includegraphics[clip, trim=0cm 0.9cm 0cm 0cm, width=0.95\columnwidth]{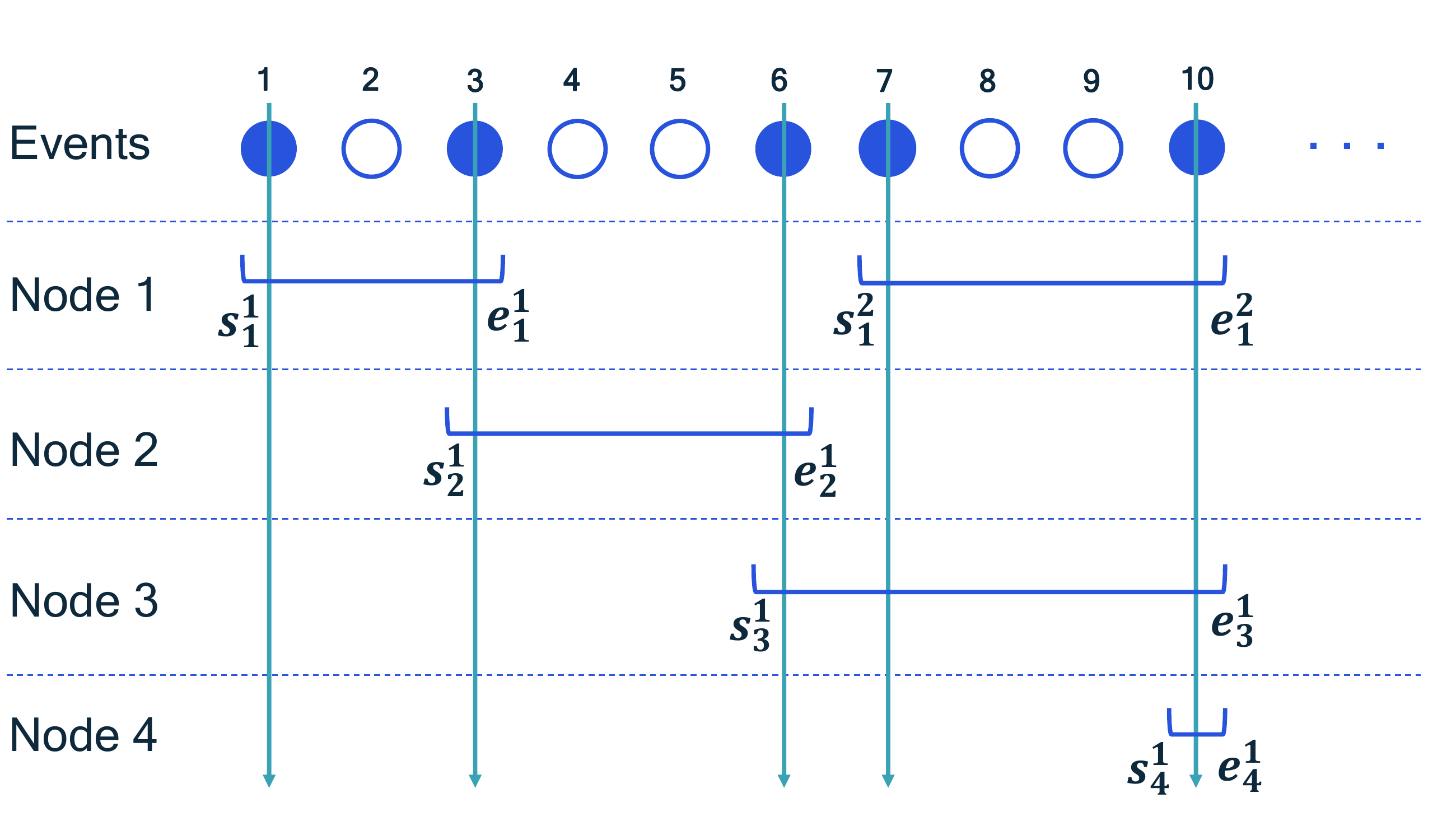}
    \vspace{-3mm}
    \caption{Visualization of the retention intervals formulation.}
    \label{fig:formulation_drawing_a_new}
\end{figure}

The concept of time in our formulation is defined through \textit{compute events}. This is one of the key components of our proposed formulation. To better illustrate this point, consider Figure \ref{fig:formulation_drawing_a_new}, which depicts our formulation for the 4-node example graph in Figure \ref{fig:4nodegraph}. In this example, we have set $C_v=2$ for all nodes $v$. Figure \ref{fig:formulation_drawing_a_new} illustrates the formulation for this graph where the vertical axis corresponds to different nodes and the horizontal axis is the time partitioned into events. The formulation as visualized in Figure \ref{fig:formulation_drawing_a_new} shows two intervals for node 1 and one interval for the other nodes. The reason for this is that the optimal solution for this example case requires only one active interval for nodes 2, 3, 4 while requiring two active intervals for node 1. We note that the starting and ending positions of the intervals in Figure \ref{fig:formulation_drawing_a_new} are obtained by solving the optimization problem. These are not known ahead of time, but rather modeled as variables in the optimization problem. Furthermore, this figure depicts circles as visual representation of potential node execution events. For example, in Figure \ref{fig:formulation_drawing_a_new}, node 1 is computed during event 1 and its output is kept in memory until event 3. The second interval for node 1 starts in event 7 indicating that node 1 is recomputed during event 7. Observe that event 1 has a memory usage of $m_1$, while event 7 has a memory usage of $m_1+m_3$, as the output of node 3 is retained in memory when node 1 is recomputed.

The total duration of the rematerialization sequence is a weighted sum over the events in which a node is being executed, where the weights are simply the actual durations of the nodes (in seconds or processor cycles). Such events are marked with a filled circle in Figure \ref{fig:formulation_drawing_a_new}. We say that the \textit{peak memory footprint} of a solution is the maximum memory usage among all events. The cyan colored arrows in Figure \ref{fig:formulation_drawing_a_new} represent the memory usage of node execution events. The peak memory footprint of the solution in the example is 3, assuming each node outputs a unit-size tensor, realized at event 10. Observe that events marked with empty circles do not contribute to either the duration or the peak memory footprint of the solution.


The primary reason why we work with this event-based definition for time is to keep the domain of the variables $s_v^i, e_v^i$ at a small size. More concretely, the CP-SAT solver \cite{ortools}, which we use to solve numerically our proposed optimization problem, accepts only integer variables as it implements Lazy Clause Generation \cite{ohrimenko2009propagation}, known to be effective for constrained scheduling problems \cite{schutt2013solving}. If we instead set the time domain to be seconds (which would require quantization) or processor cycles, then the corresponding CP formulation would require variable domains of larger sizes. The domain size of the variables in the formulation has a direct impact on the solver speed especially for CP solvers.

Note that for each interval, the first time slot (shown using a filled circle in Figure \ref{fig:formulation_drawing_a_new}) is dedicated to the computation of node $v$ while the rest of the interval indicates the retention of the output of node $v$. It follows that intervals may not start at the same time due to our system model where nodes are executed sequentially, i.e. there is no parallel computing. The empty circles in Figure \ref{fig:formulation_drawing_a_new} indicate that no interval starts at that particular time, i.e., no computation occurs. The decision of which circles will be filled or empty is a by-product of the optimization problem and not known ahead of time. Furthermore, it is important to note that a memory block of size $m_v$ is allocated during the entirety of the interval. Also, observe that since the first event of an interval indicates the compute of that node, this is when the predecessors of that node must be available in memory.

\subsection{Optimization Problem Formulation}
In addition to the start and end times of the intervals, we define the Boolean variables $a_v^i$ to model whether the $i$'th interval of node $v$ is active or inactive. This grants us the flexibility of not requiring a node $v$ to be rematerialized exactly $C_v$ times. If inactive (i.e. $a_v^i=0$), the corresponding interval does not contribute to the sum in the objective as well as the sums in the memory and precedence constraints. We assume that the duration $w_v$ and output size $m_v$ for each node are known. Using the retention intervals as the foundation of our formulation, we state the rematerialization problem as an optimization problem as follows:

\begin{align}
    \underset{s,e,a}{\mbox{minimize}} \,\quad & \sum_{v,i} w_va_v^i \label{eq:objective} \\
    \mbox{subject to} \quad & s_v^i\leq e_v^i, \, \forall v,i \label{eq:valid_interval} \\
    & e_v^i\leq s_v^{i+1}, \, \forall v, \forall i<C_v \label{eq:interval_same_node} \\
    & \sum_{v,i\, :\, s_v^i \leq t \leq e_v^i} m_v a_v^i\leq M,\, \forall t \in \mathcal{D} \label{eq:memory_constraint} \\
    & \forall (u,v)\in E,\, \forall i \in \{i:  a_v^i=1\},\, \exists j \mbox{ such that } \nonumber \\
    & \qquad \quad  a_u^j=1, s_u^j+1 \leq s_v^i\leq e_u^j  \label{eq:precedence_constraint} \\
    &s_v^i \neq s_u^j, \, \forall v,u \in \{ v,u : v\neq u\},\, \forall i,j \label{eq:alldifferent}\\
    & a_v^1 = 1,\, \forall v \label{eq:first_interval_active} \\
    & s_v^i, e_v^i \in \mathcal{D},\, a_v^i \in \{0,1\}, \, \forall v,i \,. \label{eq:a_domain}
\end{align}

The objective \eqref{eq:objective} is the total duration. Constraints \eqref{eq:valid_interval} enforce that the end time of each interval comes after its start time. Constraints \eqref{eq:interval_same_node} ensure that intervals of the same node do not overlap. The memory budget constraints are given in \eqref{eq:memory_constraint}, where the parameter $t$ represents time. The precedence constraints \eqref{eq:precedence_constraint} enforce that there exists an overlapping and active interval for all predecessors of $v$ at all start times $s_v^i$ of $v$. Observe that the memory and precedence constraints are nonlinear constraints. We describe in detail how the \textit{cumulative constraint} and the \textit{reservoir constraint} from constraint programming are used to model these nonlinear constraints in the next subsection. The constraints in \eqref{eq:first_interval_active} are intended to make the first retention interval for every node active. The constraints \eqref{eq:alldifferent} ensure that the starting times of the intervals are all different, i.e. compute events do not overlap. 

The domain of the variables $s_v^i, e_v^i$ is as follows
\begin{align}
    \mathcal{D} = \{1,2,\dots,\sum_{v}C_v\} \,,
\end{align}
where $\sum_{v}C_v$ is equal to the number of intervals, which is upper bounded by $|\mathcal{D}| \leq n \max_v{C_v}$. Observe that we may not have more than $\sum_{v}C_v$ filled circles in Figure \ref{fig:formulation_drawing_a_new}.

\subsection{Memory and Precedence Constraints using CP}

\textbf{Memory constraints:} The set of constraints in \eqref{eq:memory_constraint} enforces that the resource usage never exceeds the local memory budget $M$ at any time $t$. Note that the constraint \eqref{eq:memory_constraint} is not linear in the variables $s_v^i, e_v^i$. This type of inequality can be modeled using the CP constraint $\verb|cumulative|$.

The cyan colored arrows in Figure \ref{fig:formulation_drawing_a_new} are intended to visually represent how we model the memory constraint. Each vertical arrow can be viewed as a separate inequality constraint. In particular, the sum of output sizes $m_v$ for each interval an arrow intersects is less than or equal to $M$. Recall that it is not needed to have an arrow for each circle in Figure \ref{fig:formulation_drawing_a_new} since the memory use could not increase during the empty circles. Hence, we only need to consider the filled circles. However, we do not know which circles are to be filled ahead of time. This is readily addressed by the cumulative modeling which considers only the beginning of intervals for memory calculation.


We have used the function \texttt{AddCumulative} from CP-SAT \cite{ortools} for the memory constraints. This function requires intervals, demands, and capacity as input arguments. Intervals are the retention intervals from our formulation defined by the start and end variables $s_v^i, e_v^i$ and considered only when $a_v^i=1$. The demand for each interval is the output size $m_v$ for the corresponding node $v$. Finally, the capacity is simply the memory budget $M$. The pseudo-code representation of this constraint is as follows:
%
%
%
$$\verb|cumulative| (\{s_v^i, e_v^i, a_v^i, m_v\}_{v\in V, i\in \{1..C_v\}}, M) \,.$$

\textbf{Precedence constraints:} We recall that prior to the execution of each compute task, the outputs of all of its predecessors must be available in memory. This could be written as the constraints in \eqref{eq:precedence_constraint}. Note that these are nonlinear constraints in the variables. We employ the $\verb|reservoir|$ constraint from constraint programming to model the data dependencies. We view each predecessor as a resource whose level needs be maintained such that it does not go below $0$ while its successor is being executed. 

Let us define the resource level change events $f(\cdot)$ for each edge $(u,v) \in E$ and $i \in \{1,\dots,C_v\}$:
\begin{align} \label{eq:level_change_events}
    &f(s_v^i) = -1 \nonumber \\
    &f(s_v^i+1) = 1 \nonumber \\
    &f(s_u^1) = 1, f(s_u^2) = 1, \dots, f(s_u^{C_u}) = 1 \nonumber \\
    &f(e_u^1) = -1, f(e_u^2) = -1,\dots, f(e_u^{C_u}) = -1 \,.
\end{align}
The level change function $f(\cdot)$ returns $0$ at every other point in time except for those in \eqref{eq:level_change_events}. The function \texttt{AddReservoirConstraintWithActive} from CP-SAT takes the following input parameters: times, demands, actives, minimum level. Times and demands are as defined in \eqref{eq:level_change_events}. Actives are the Boolean variables $a_v^i$ and the minimum level is set to $0$. This constraint ensures that for node $v$, one of the intervals for each of its predecessors $u$ overlaps with the starting time of the node $v$. 
%

Finally, the nonlinear constraints \eqref{eq:alldifferent} could be addressed using the \texttt{alldifferent} constraint from CP.

\subsection{Complexity Reduction by Enforcing an Initial Topological Ordering} \label{sec:enforce_topo}
The proposed framework offers the flexibility of allowing solutions that are not limited to a predetermined topological ordering. However, one could enforce an input topological ordering to reduce the size of the search space and in turn reduce the solve time. We expand upon this point in this subsection.

\begin{figure}
    \centering
    \includegraphics[clip, trim=0cm 0.9cm 0cm 0cm, width=0.95\columnwidth]{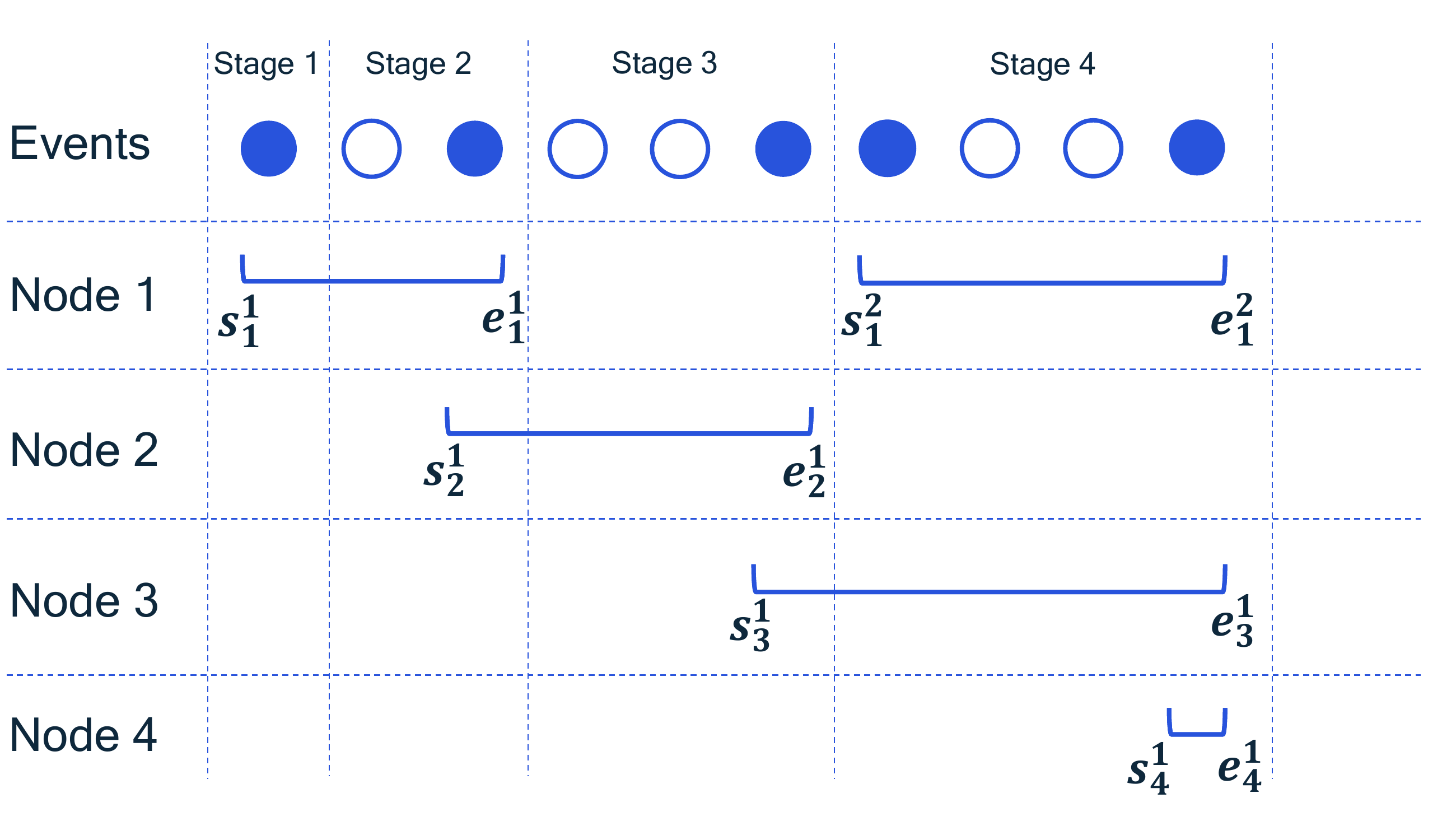}
    \vspace{-3mm}
    \caption{Illustration of events grouped into stages.}
    \label{fig:formulation_drawing_b_new}
\end{figure}

We start by considering a new domain for the integer variables $s_v^i, e_v^i$:
\begin{align}
    \mathcal{D} = \Big\{1,2,\dots,\frac{n(n+1)}{2}\Big\} \,,
\end{align}
where the domain size is now $O(n^2)$ rather than $O(n)$. We now introduce the concept of \textit{stages} into our formulation, which has been previously employed in \cite{jain2020checkmate}. The $j$'th stage contains $j$ events; this is depicted in Figure \ref{fig:formulation_drawing_b_new}. Given a topological ordering of the nodes, the first stage enforces the computation of the first node in the order. In subsequent stages, the $j$'th node in the input topological order is enforced to be computed in the last event of stage $j$, while the previous events in stage $j$ allow nodes $1,\dots,(j-1)$ to be computed. In particular, we enforce that node $j$ could be computed only in event $j$ of a stage. Since this restricts the starting times of each interval to specific events, we no longer need to explicitly include the constraint \eqref{eq:alldifferent}.

Note that the start time for the first retention interval for each node is no longer a variable but a fixed value. More precisely, for node $v$, the value of $s_v^1$ is equal to $j(j+1)/2$, where $j$ is the index of node $v$ in the input topological ordering.


\subsection{Optimization in Two Phases}
Our approach consists of two phases. In Phase 1, we solve a variant of the optimization problem in \eqref{eq:objective} - \eqref{eq:a_domain} to find a memory feasible solution. This solution is then used as initialization for Phase 2, which is the optimization problem stated in \eqref{eq:objective} - \eqref{eq:a_domain}. The problem in Phase 1 has the objective
\begin{align}
    \underset{s, e, a, M_{var}}{\mbox{minimize}} \quad \max(M_{var}, M)
\end{align}
where we introduce the variable $M_{var} \in \mathbb{R}$ for the peak memory footprint. We modify the memory constraint in \eqref{eq:memory_constraint} as follows for Phase 1:
\begin{align}
    \sum_{v,i\, :\, s_v^i \leq t \leq e_v^i} m_v a_v^i\leq M_{var},\, \forall t \in \mathcal{D}\,.
\end{align}
The goal of Phase 1 is to arrive at an intermediate solution, whose peak memory footprint is below the local memory target $M$. We address this by considering the maximum of the peak memory variable $M_{var}$ and memory budget $M$ as the objective. This objective could be linearized by introducing an auxiliary variable $\tau \in \mathbb{R}$ and then considering: ${\mbox{minimize}} \,\, \tau$ subject to $\tau \geq M_{var}$ and $\tau \geq M$. The other constraints of the optimization problem are unchanged. 

Note that any topological ordering of the graph $G$ provides a trivial feasible solution to the problem in Phase 1 since it does not have a hard memory constraint. For the rematerialization problem itself, however, there is no easy solution that could be constructed and provided as an initial solution. Phase 1 is intended to fill this gap.

\subsection{Overview of Formulations}
Table \ref{table:overview_formulations} provides a complexity comparison for our approach and the approach of \cite{jain2020checkmate}.
When comparing the number of constraints for different formulations, it is important to keep in mind that the constraints in the MILP are all linear constraints while the constraints of the CP could be linear or nonlinear.

\begin{table*}[hbt!]
\caption{Overview of formulation complexities for \textsc{Checkmate} and our proposed method.
}
\vspace{3mm}
\label{table:overview_formulations}
\centering
\begin{tabular}{||c c c c c||} 
 \hline
 Formulation & \# Bool. vars & \# integer vars &  Size of search space & \# constraints\\ [0.5ex] 
 \hline\hline
 \textsc{Checkmate} - MILP & $O(n^2+nm)$ & - & $O(2^{n^2 + nm})$ & $O(n^2+nm)$ \\ 
 \hline
 \textsc{Moccasin} - CP & $O(Cn)$ & $O(Cn)$ (domain size = $O(n)$) & $O(2^{Cn} +  n^{Cn})$ & $O(Cm)$\\
 \hline
\end{tabular}
\end{table*}

\begin{remark}
The optimization problem in \eqref{eq:objective} - \eqref{eq:a_domain} is specified using integer variables $s_v^i$ and $e_v^i$ while the discrete variables of the MILP in \cite{jain2020checkmate} are all Boolean. The fact that our formulation is stated using integer variables as opposed to Boolean variables is only for convenience. For a more direct comparison of variable counts, it is straightforward to rewrite this optimization problem by representing the integer variables as Boolean sequences, in which case the number of Boolean decision variables will be $O(Cn\log n)$ (taking $C=\max_v C_v$) as opposed to $O(n^2)$ Boolean variables in the \textsc{Checkmate} formulation.
\end{remark}

Note that the size of search space in Table \ref{table:overview_formulations} for \textsc{Moccasin} is dominated by the term $n^{Cn}$. For the version of \textsc{Moccasin} with Boolean decision variables, the search space size would scale with $2^{Cn\log n}$, which is the same as $n^{Cn}$. This is smaller than the size of the search space for \textsc{Checkmate}, which is $2^{n^2+nm}$.

\section{Numerical Results} \label{sec:numerical_results}

\subsection{Graphs for Evaluation}
The graphs we use for evaluation of the method presented in this paper come from several sources.  Additional descriptions can be found in Appendix \ref{sec:app_graphs}.

\textbf{Checkmate Graphs:} 
We use graphs from the checkmate repository \cite{checkmate_repo}. They represent the single-batch training computation graphs for a selected set of neural networks. Use of these graphs allows us to compare directly to results presented in that work.

\textbf{Random Layered Graphs:} 
We leverage the synthetically constructed random layered graphs introduced in (\citealp[Appendix A]{topoformer}) as examples of graphs with complex interconnect topology.

\textbf{Real-world Graphs:}
While our synthetic and standard, public-domain graphs are valuable for experimentation, we also see value in presenting results obtained from neural computation graphs used for commercial development of artificial intelligence hardware and software products.  We sample a set of representative graphs that have diverse architectures and size.

\subsection{Experimental Setup}
We use the open source CP-SAT solver from Google OR-Tools \cite{ortools} to solve the CP. 
For the comparisons against \textsc{Checkmate}, we have used the implementation provided in the codebase for \cite{jain2020checkmate}, which utilizes Gurobi \cite{gurobi} and CVXPY \cite{diamond2016cvxpy, agrawal2018rewriting}. We have run all of the numerical experiments on a 16-CPU core workstation with a 32 GB of RAM. Furthermore, we note that once a solution to the optimization problem is obtained, the corresponding sequence of tasks can be executed on any hardware, CPU or GPU. We note that this is due to the generality of our approach where we do not make any assumptions on the computing unit that will be used to execute the computations.

\subsection{Experimental Results}

\begin{figure*}[hbt!]
    \centering
        
    \begin{minipage}{.24\linewidth}
        \centering
        \includegraphics[width=\columnwidth]{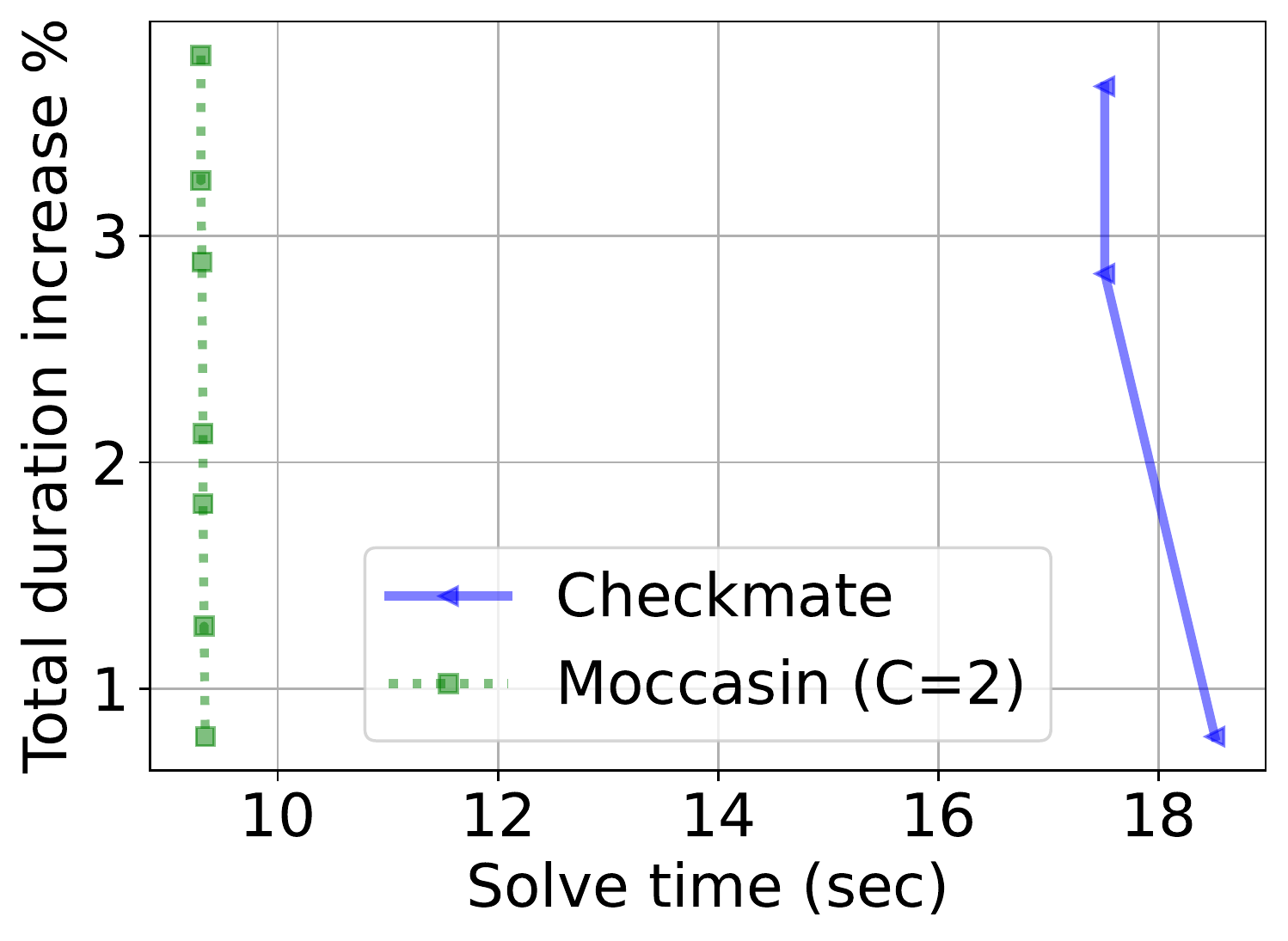}
         \centerline{\scriptsize $G_1, M=41687$}\medskip
    \end{minipage}%
    \begin{minipage}{.24\linewidth}
        \centering
        \includegraphics[width=\columnwidth]{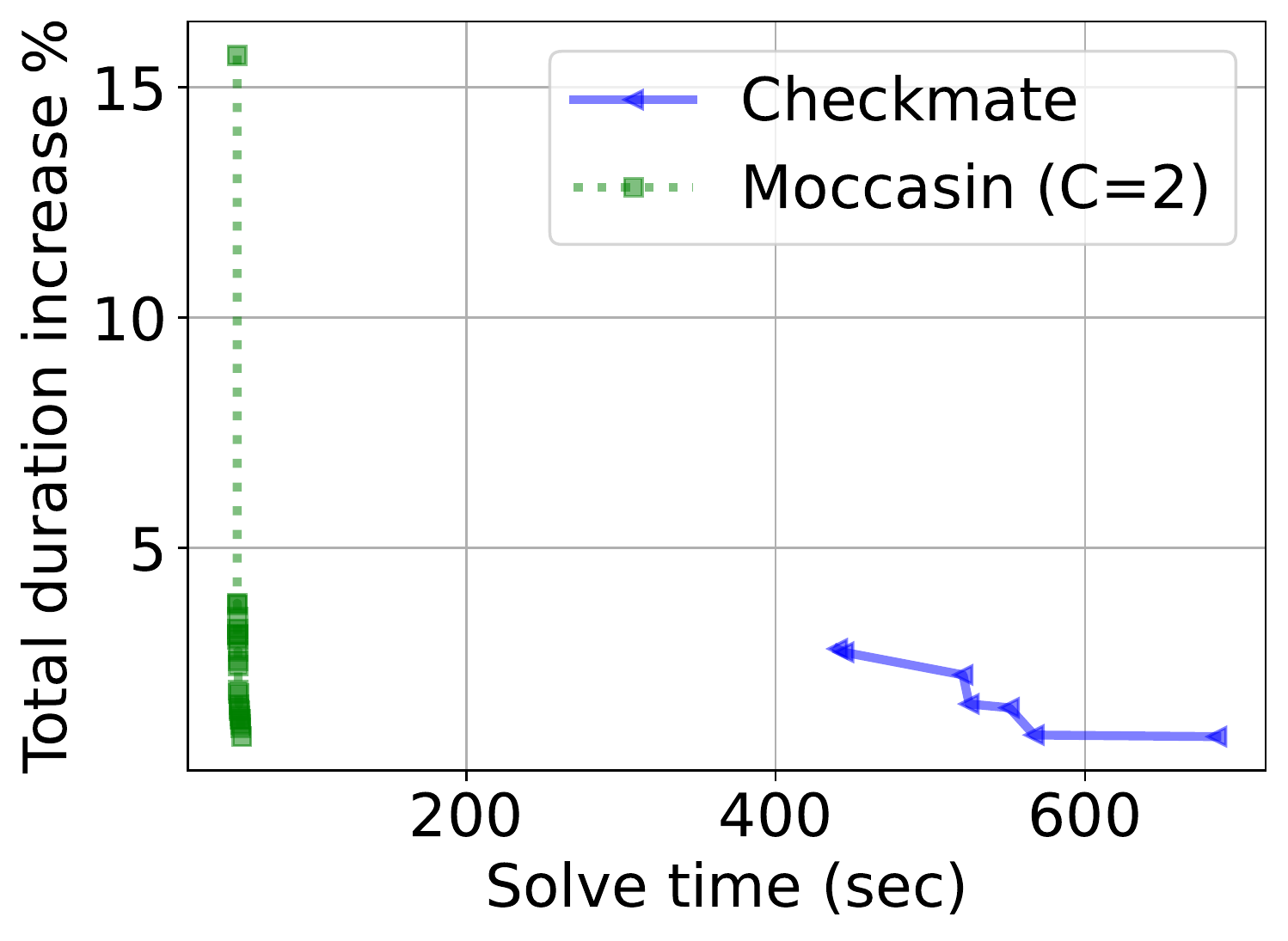}
         \centerline{\scriptsize $G_2, M=132156$}\medskip
    \end{minipage}
    \begin{minipage}{.24\linewidth}
        \centering
        \includegraphics[width=\columnwidth]{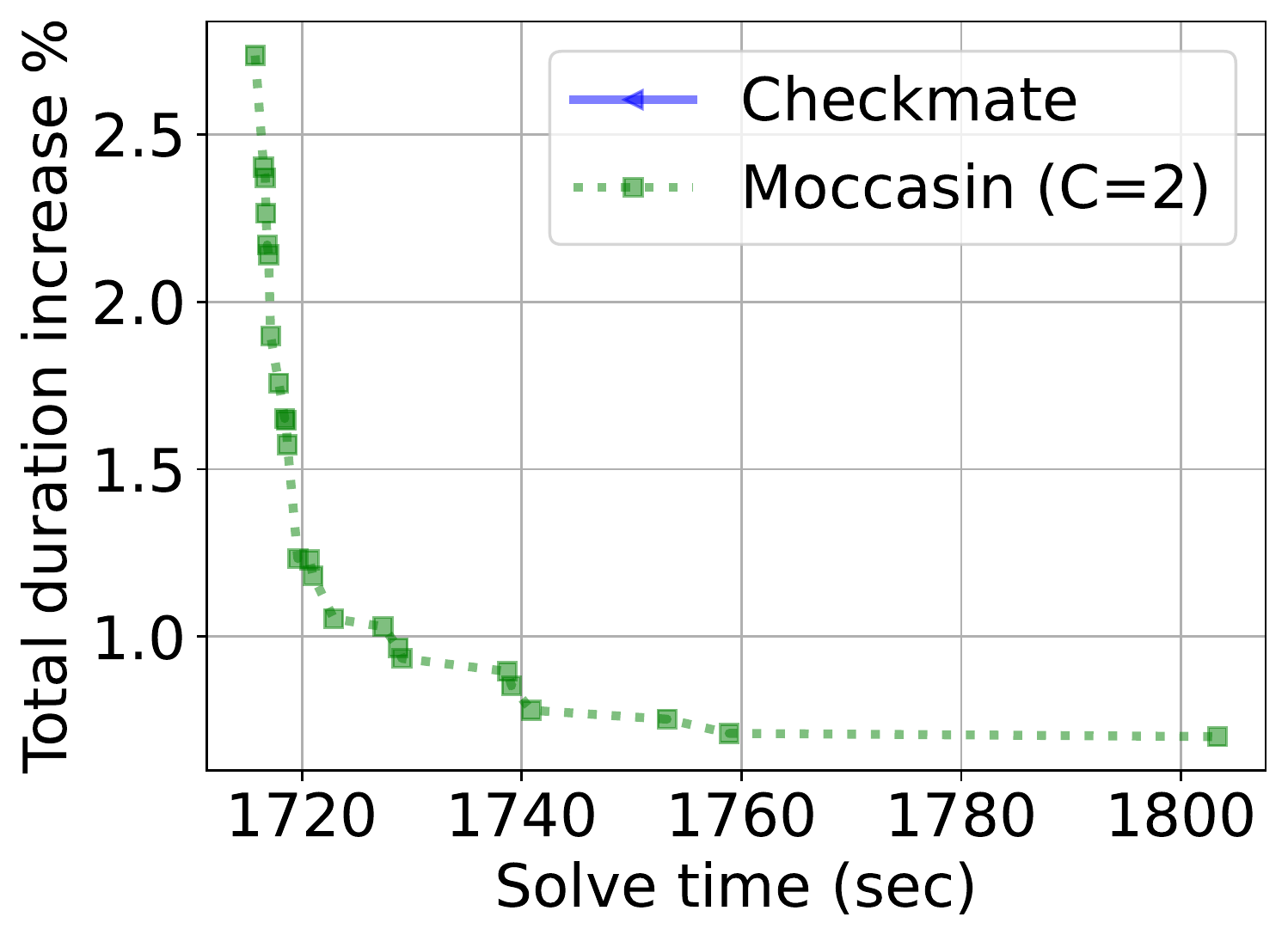}
         \centerline{\scriptsize $G_3, M=255995$}\medskip
    \end{minipage}
    \begin{minipage}{.24\linewidth}
        \centering
        \includegraphics[width=\columnwidth]{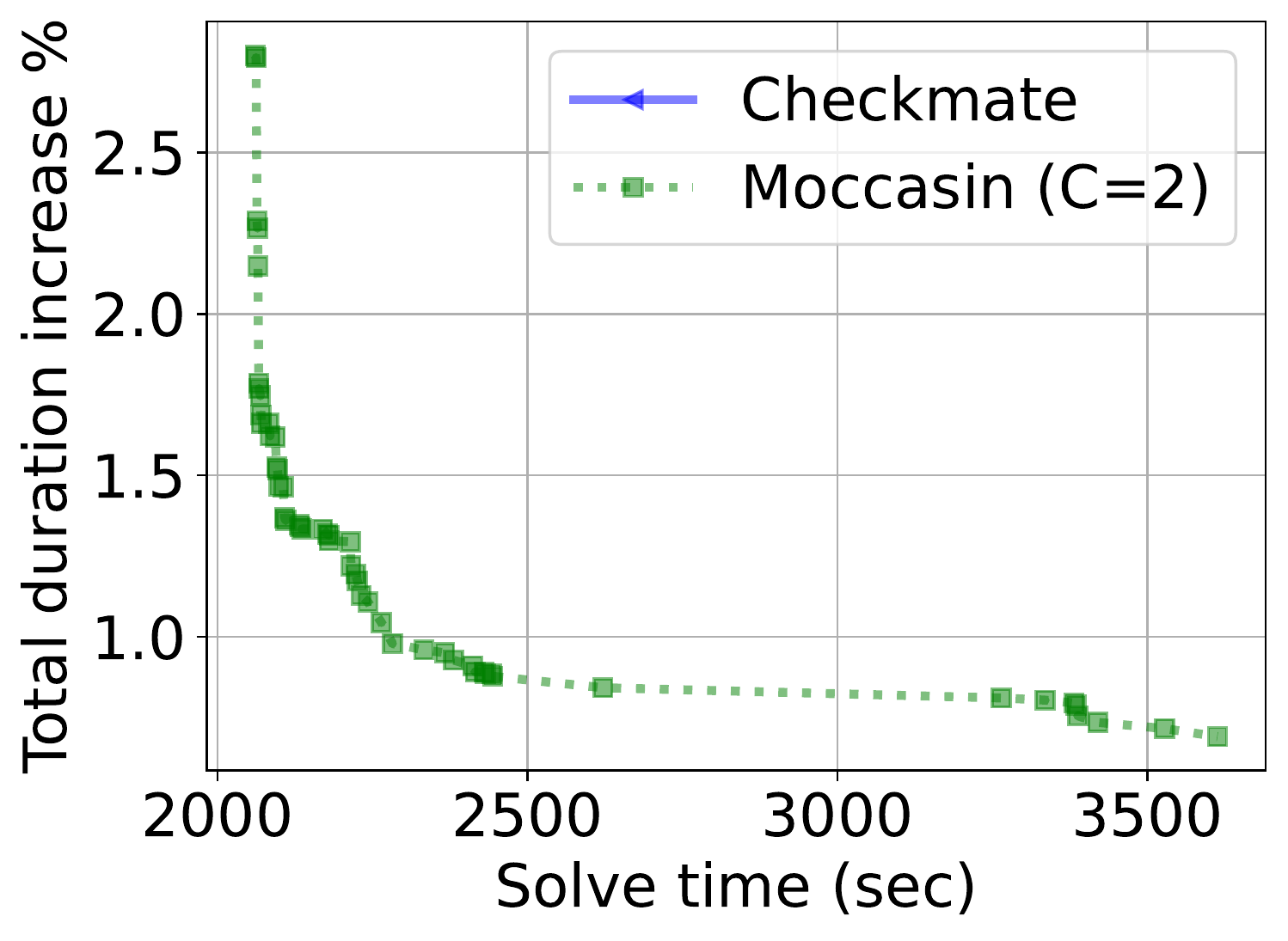}
         \centerline{\scriptsize $G_4, M=547757$}\medskip
    \end{minipage}
        \begin{minipage}{.24\linewidth}
        \centering
        \includegraphics[width=\columnwidth]{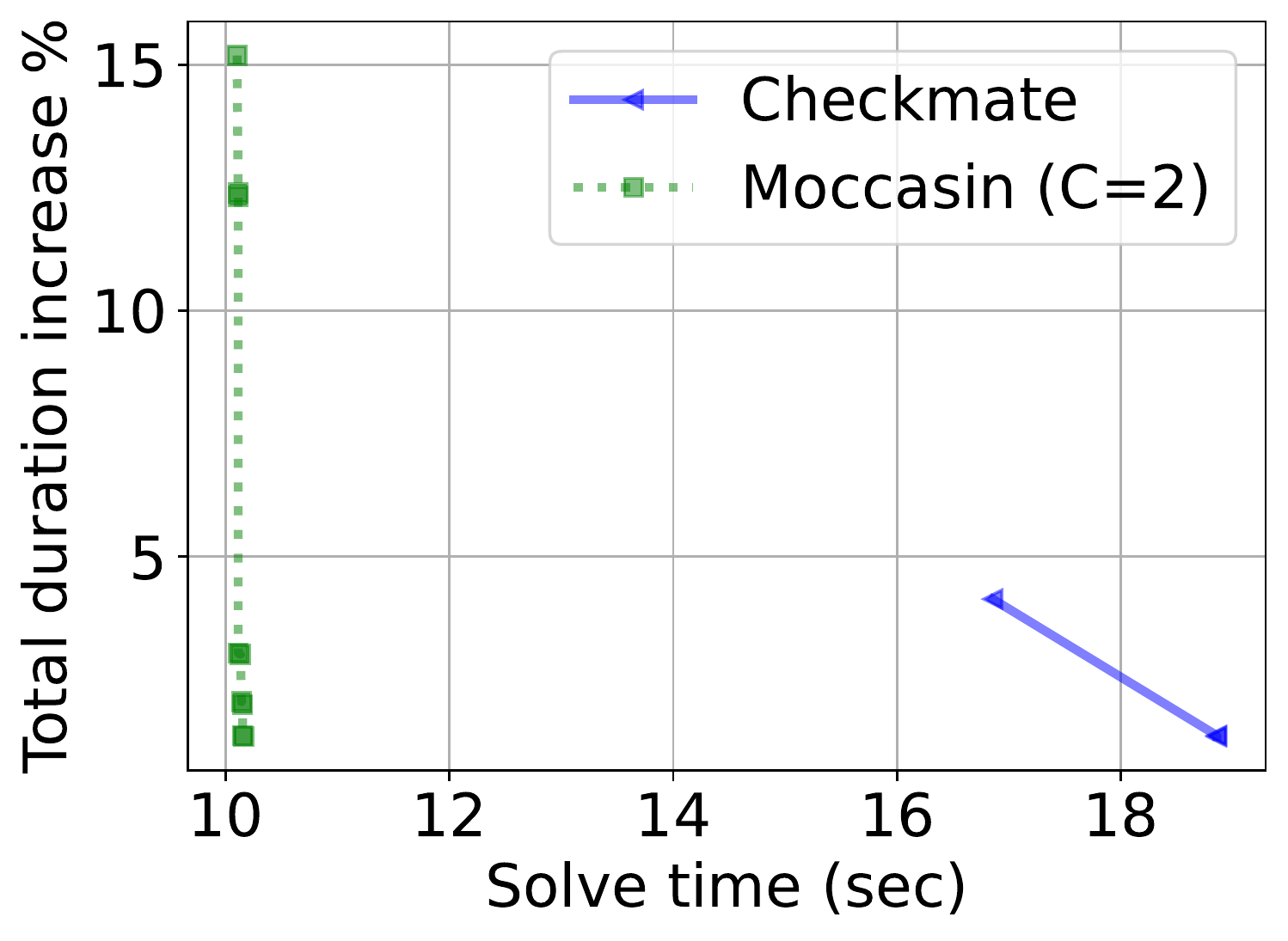}
         \centerline{\scriptsize $G_1, M=39371$}\medskip
    \end{minipage}%
    \begin{minipage}{.24\linewidth}
        \centering
        \includegraphics[width=\columnwidth]{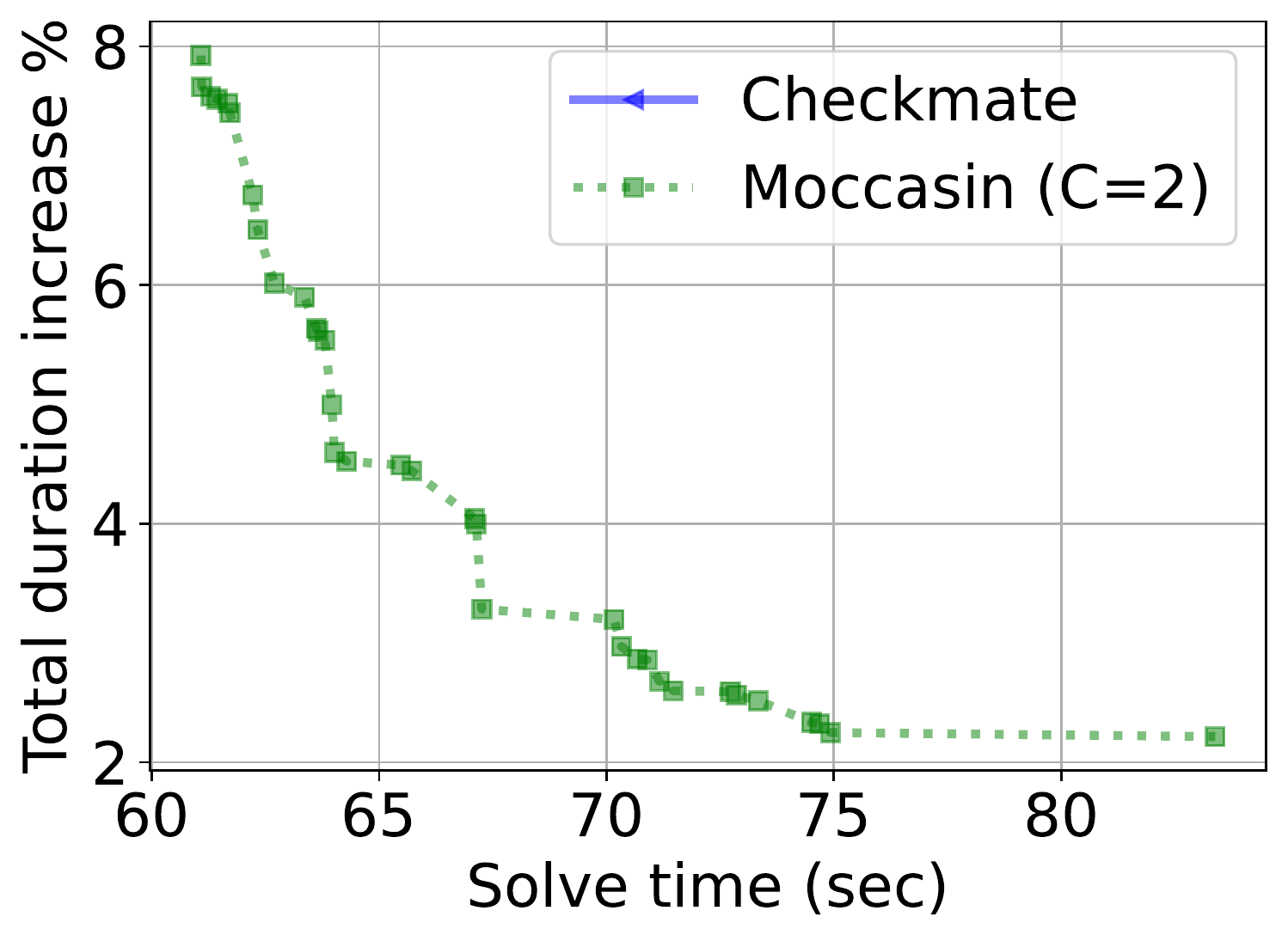}
         \centerline{\scriptsize $G_2, M=124814$}\medskip
    \end{minipage}
    \begin{minipage}{.24\linewidth}
        \centering
        \includegraphics[width=\columnwidth]{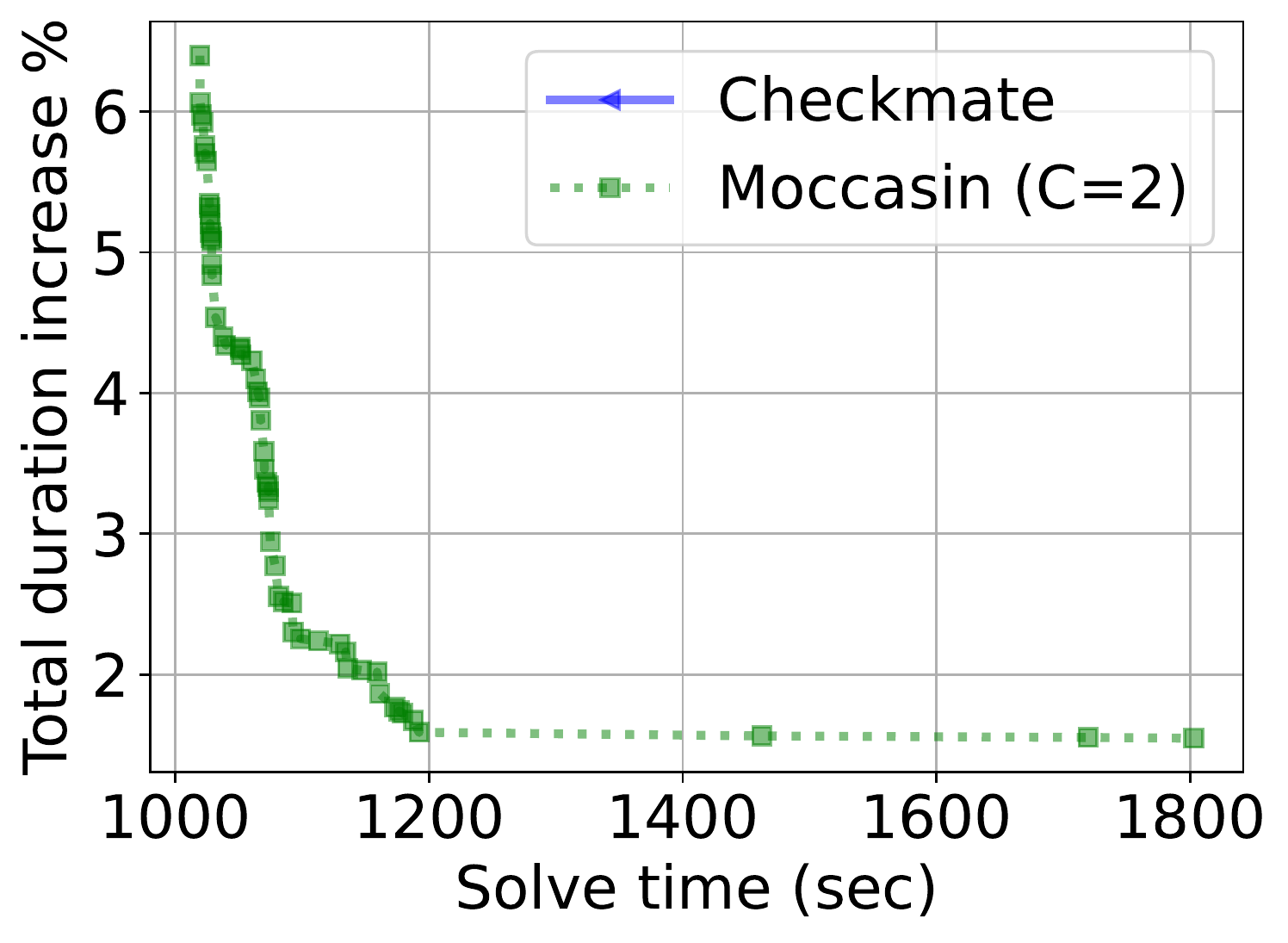}
         \centerline{\scriptsize $G_3, M=241773$}\medskip
    \end{minipage}
    \begin{minipage}{.24\linewidth}
        \centering
        \includegraphics[width=\columnwidth]{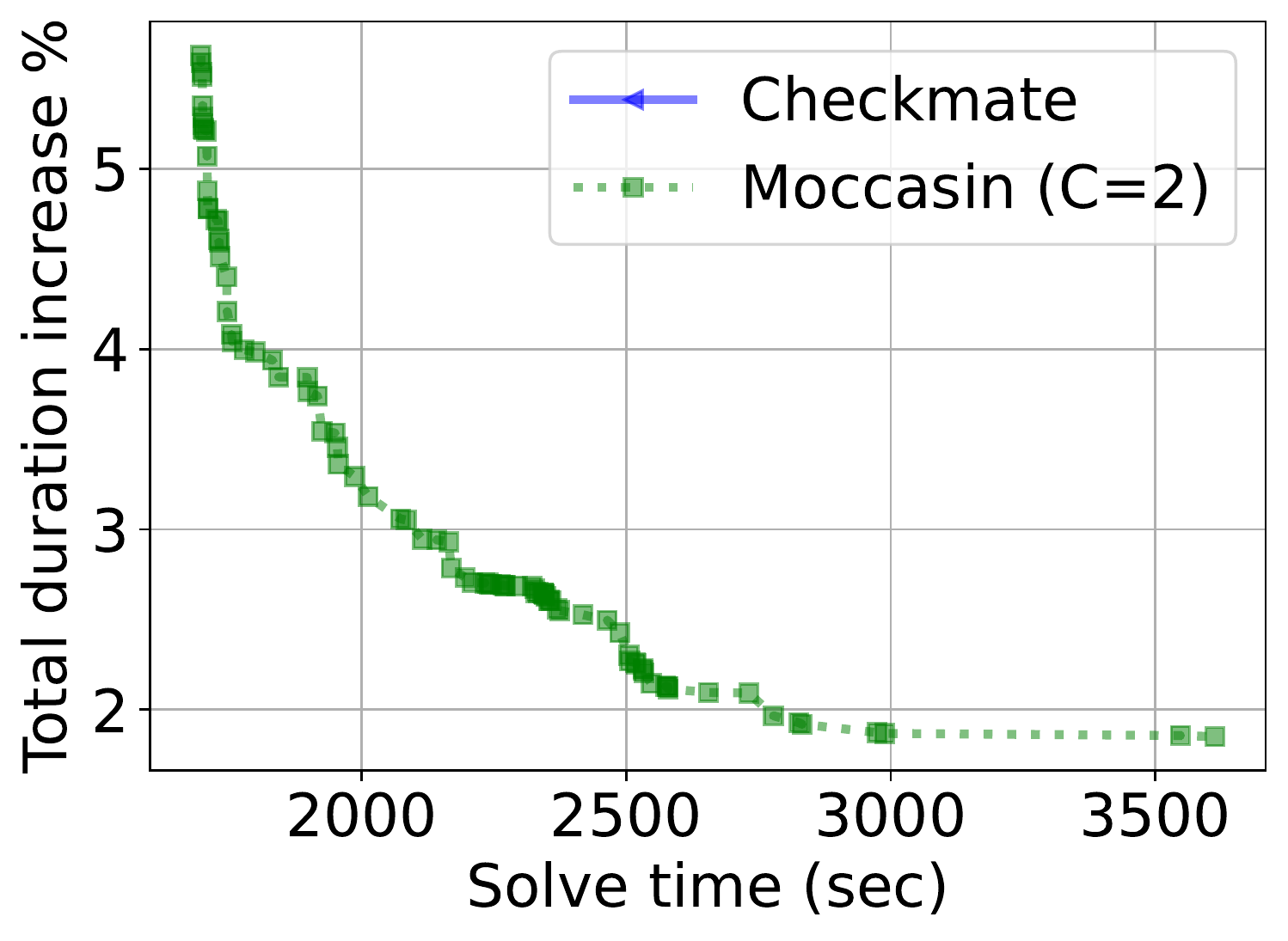}
         \centerline{\scriptsize $G_4, M=517326$}\medskip
    \end{minipage}
    \begin{minipage}{.24\linewidth}
        \centering
        \includegraphics[width=\columnwidth]{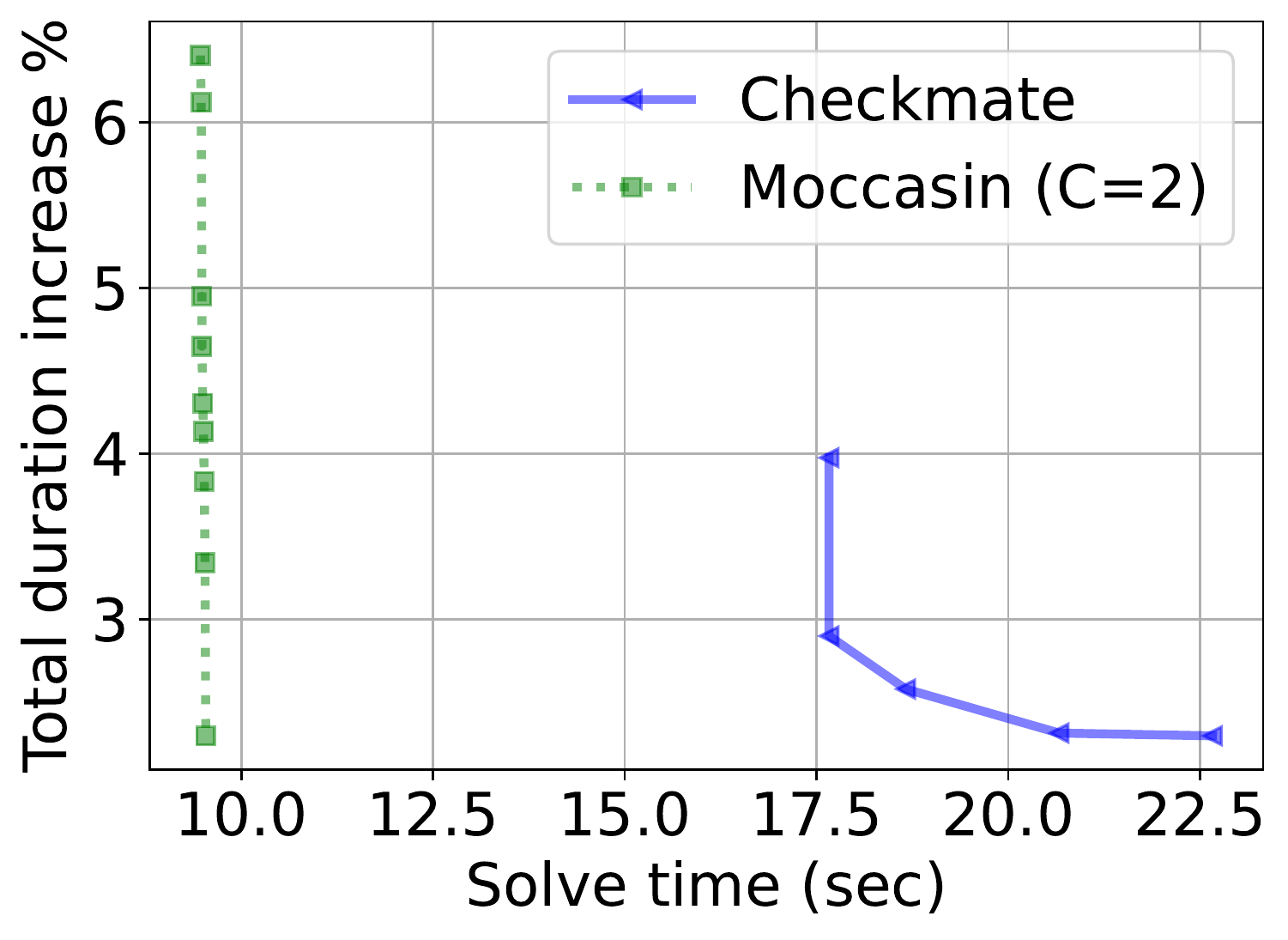}
        \centerline{\scriptsize $G_1, M=37055$}\medskip
    \end{minipage}%
    \begin{minipage}{.24\linewidth}
        \centering
        \includegraphics[width=\columnwidth]{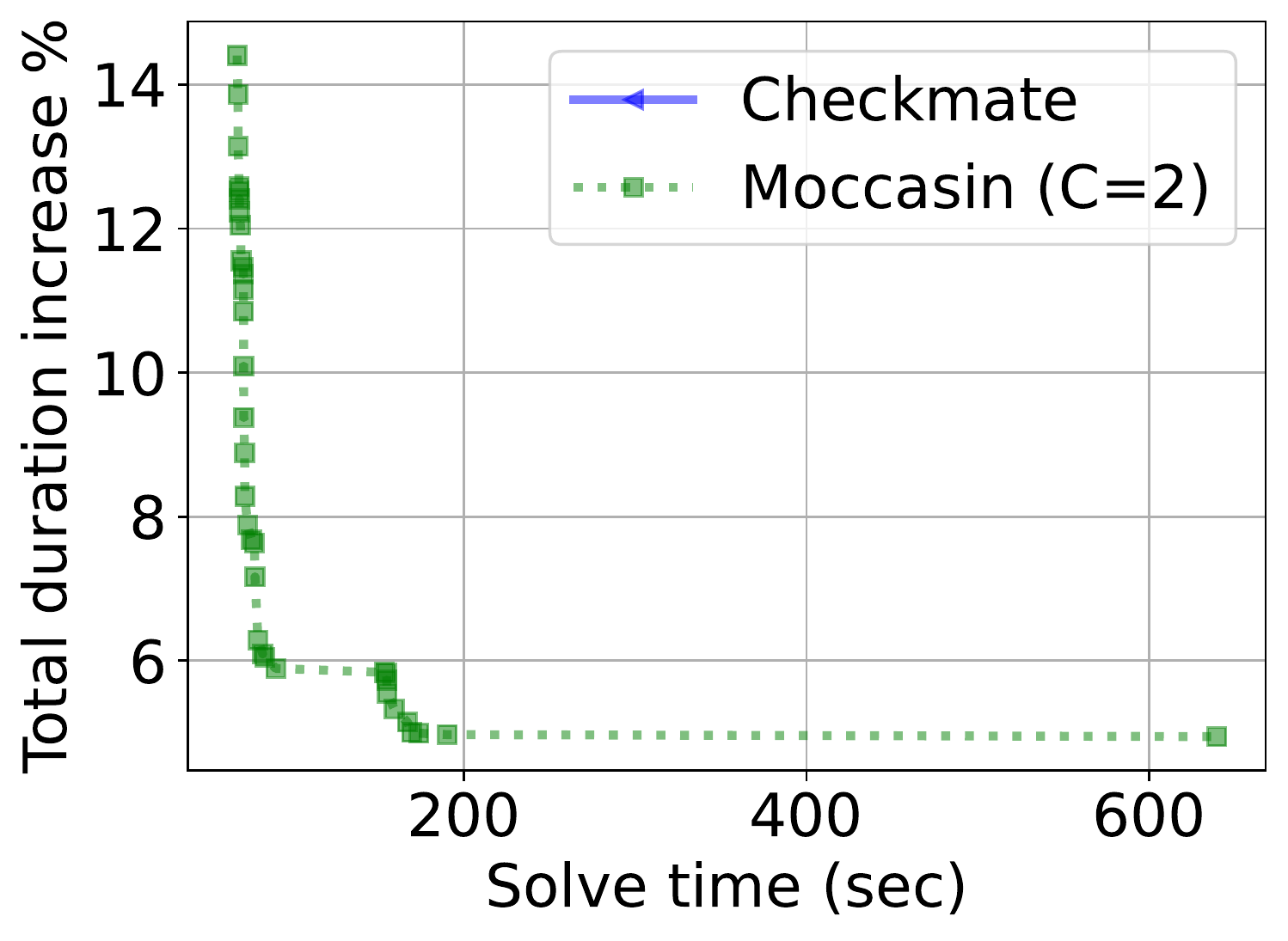}
         \centerline{\scriptsize $G_2, M=117472$}\medskip
    \end{minipage}
    \begin{minipage}{.24\linewidth}
        \centering
        \includegraphics[width=\columnwidth]{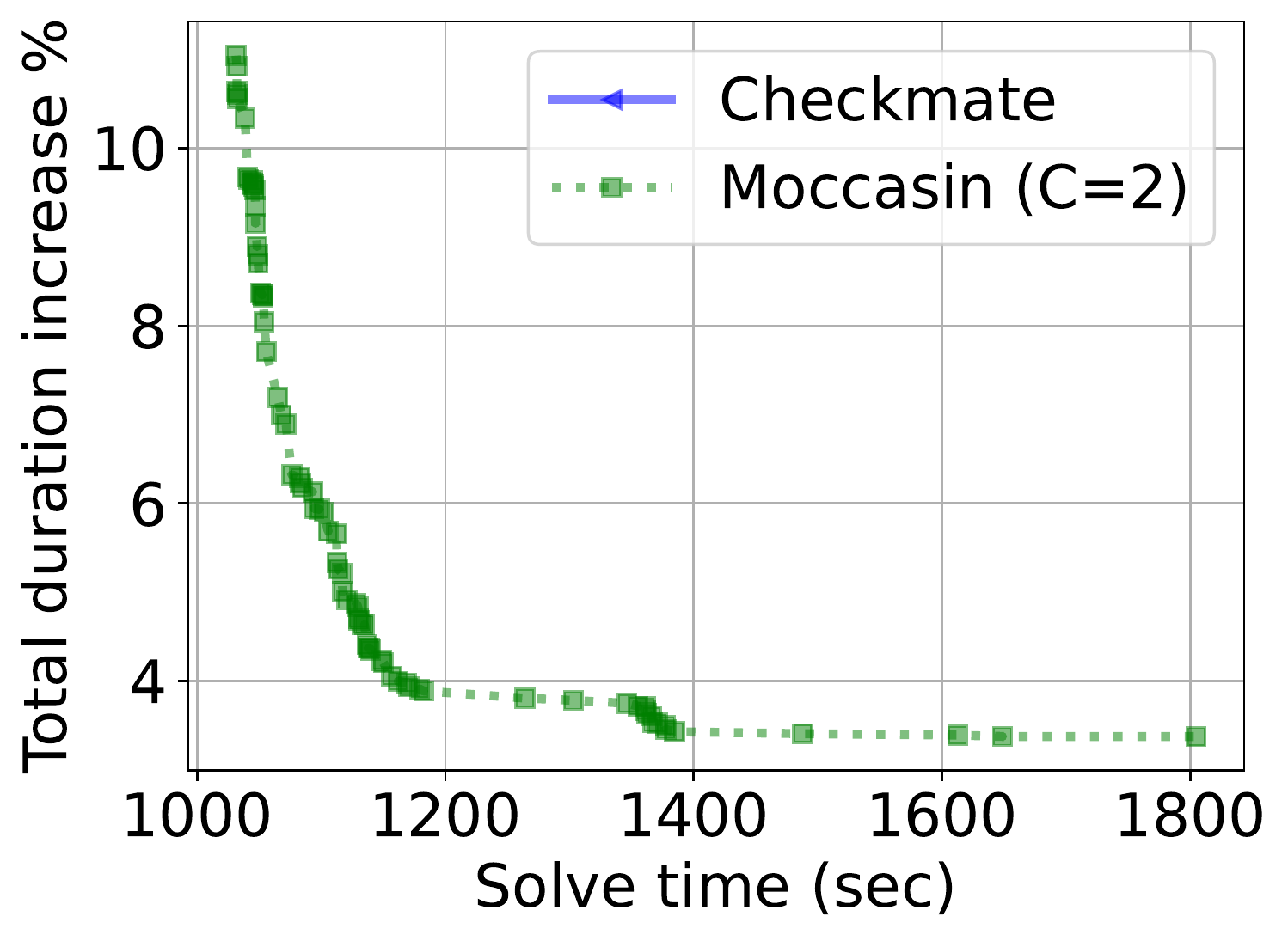}
         \centerline{\scriptsize $G_3, M=227551$}\medskip
    \end{minipage}
    \begin{minipage}{.24\linewidth}
        \centering
        \includegraphics[width=\columnwidth]{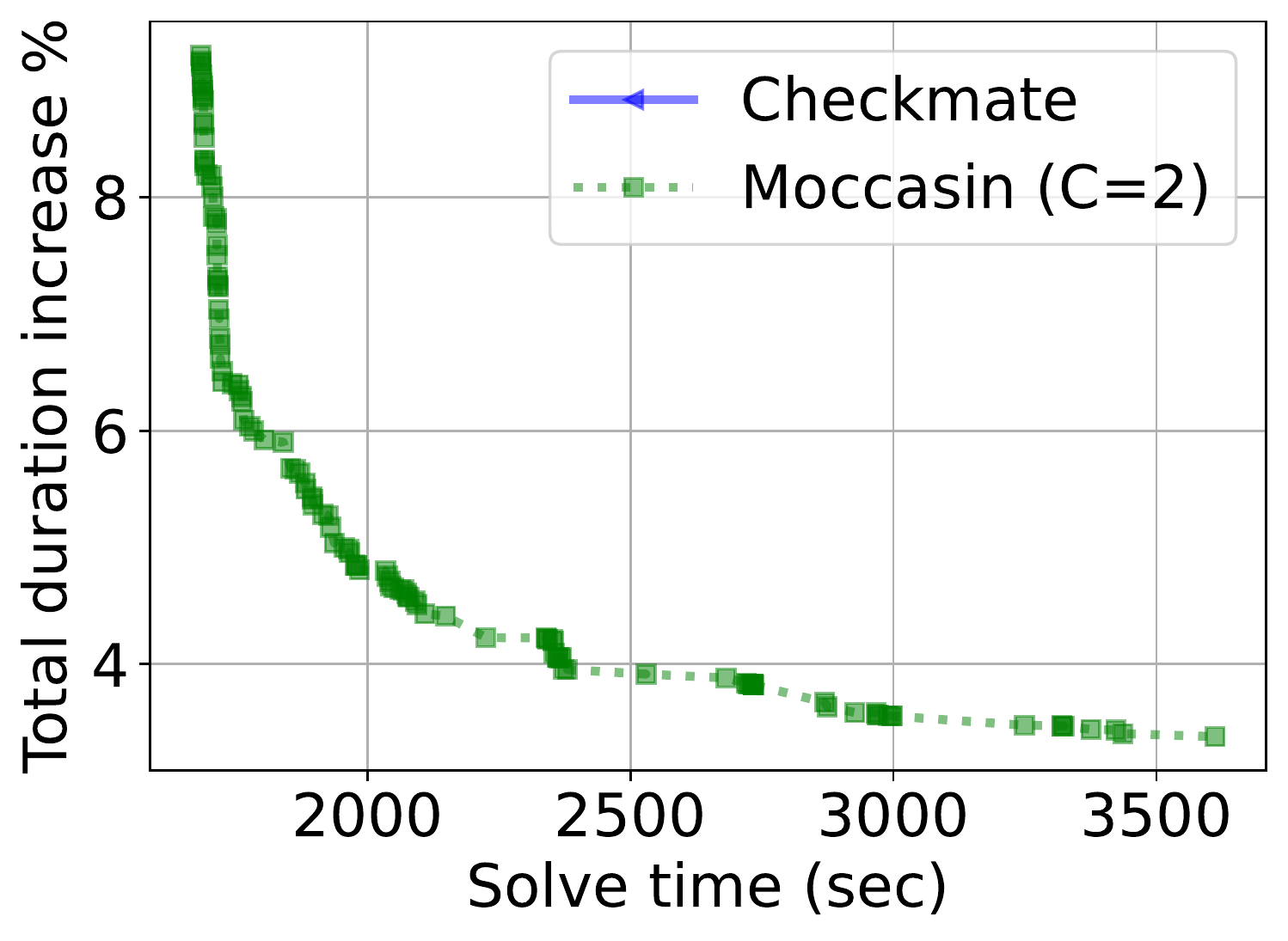}
         \centerline{\scriptsize $G_4, M=486895$}\medskip
    \end{minipage}
    \begin{minipage}{.24\linewidth}
        \centering
        \includegraphics[width=\columnwidth]{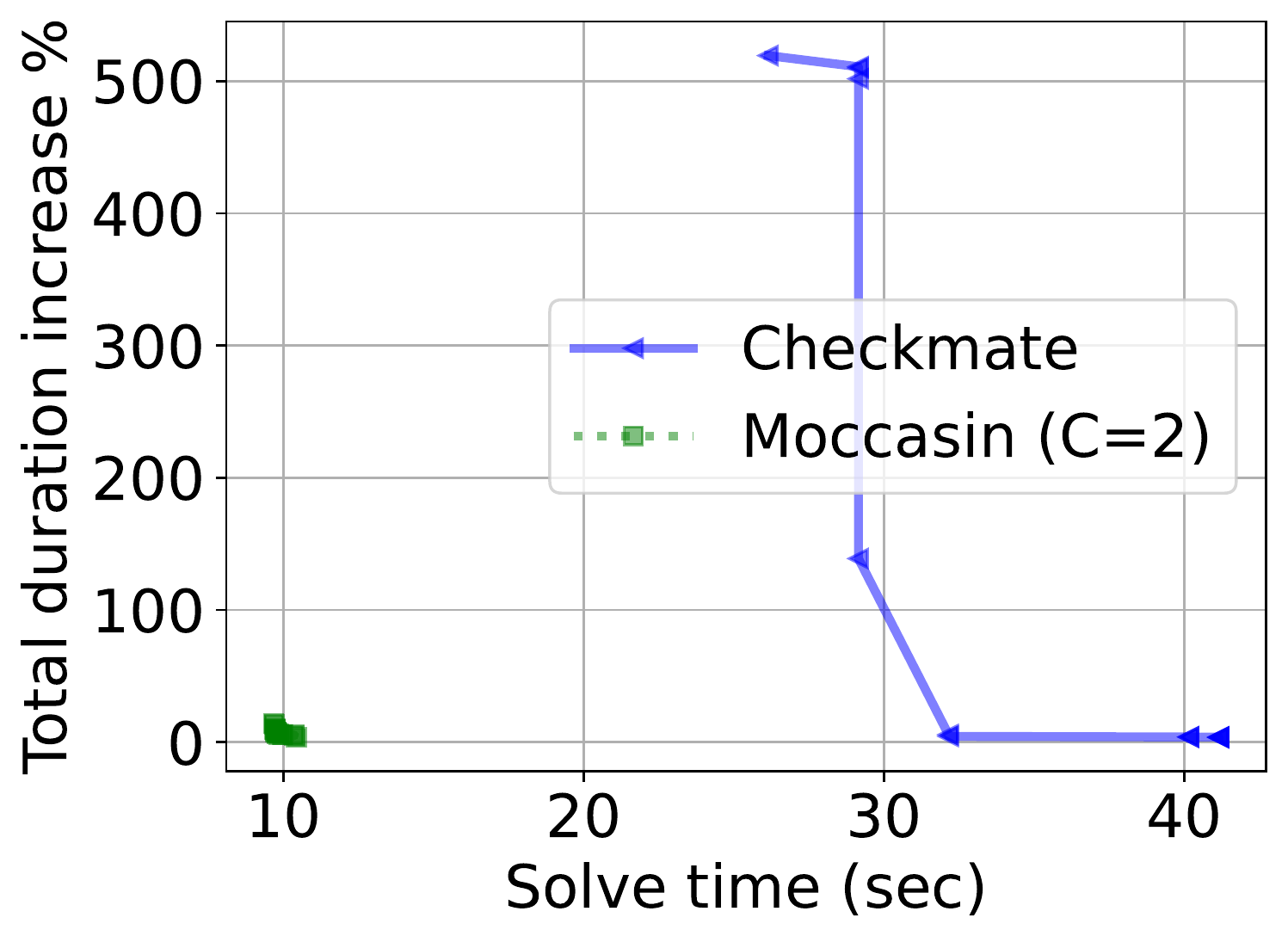}
        \centerline{\scriptsize $G_1, M=34739$}\medskip
    \end{minipage}%
    \begin{minipage}{.24\linewidth}
        \centering
        \includegraphics[width=\columnwidth]{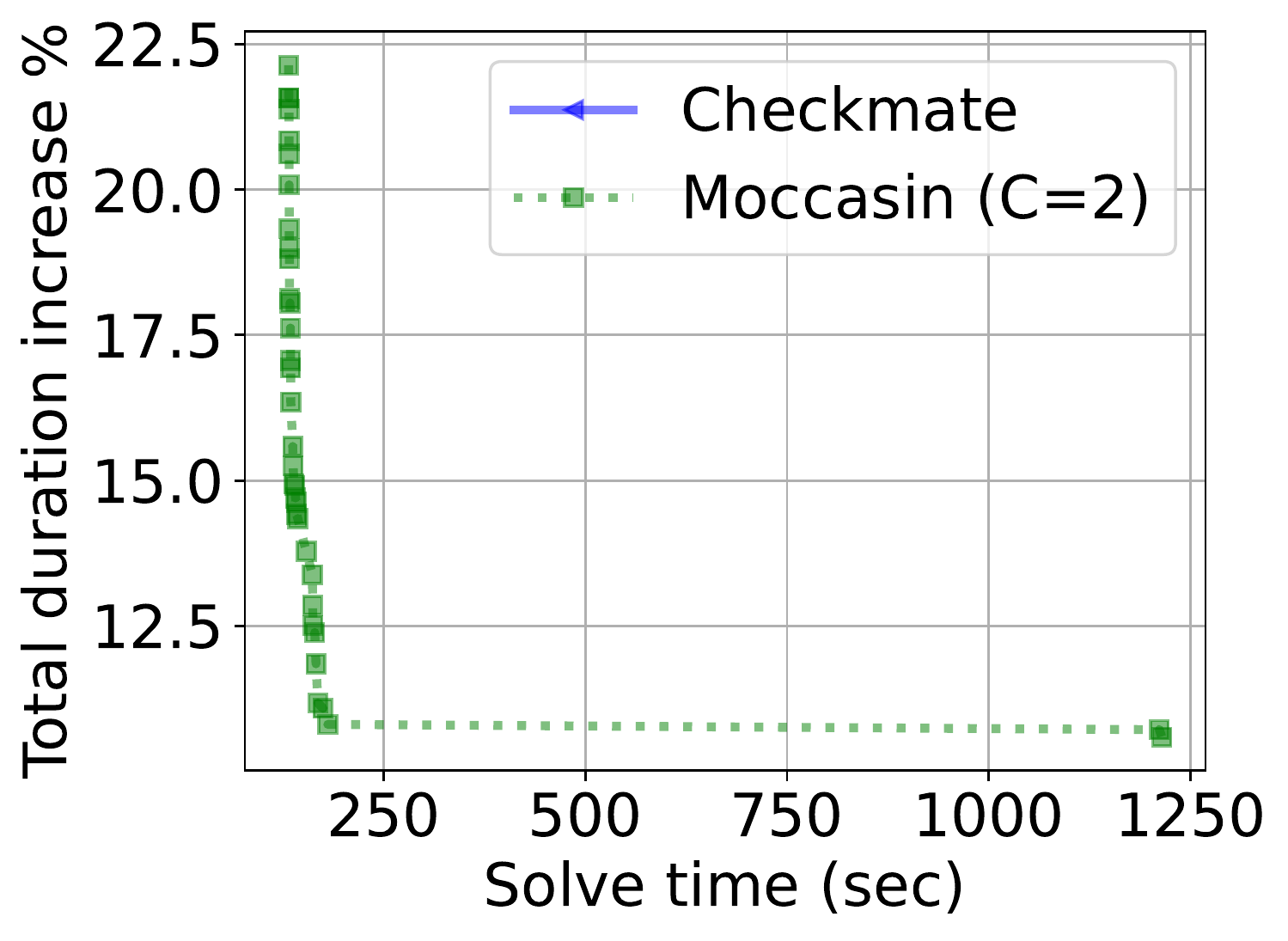}
        \centerline{\scriptsize $G_2, M=111598$}\medskip
    \end{minipage}
    \begin{minipage}{.24\linewidth}
        \centering
        \includegraphics[width=\columnwidth]{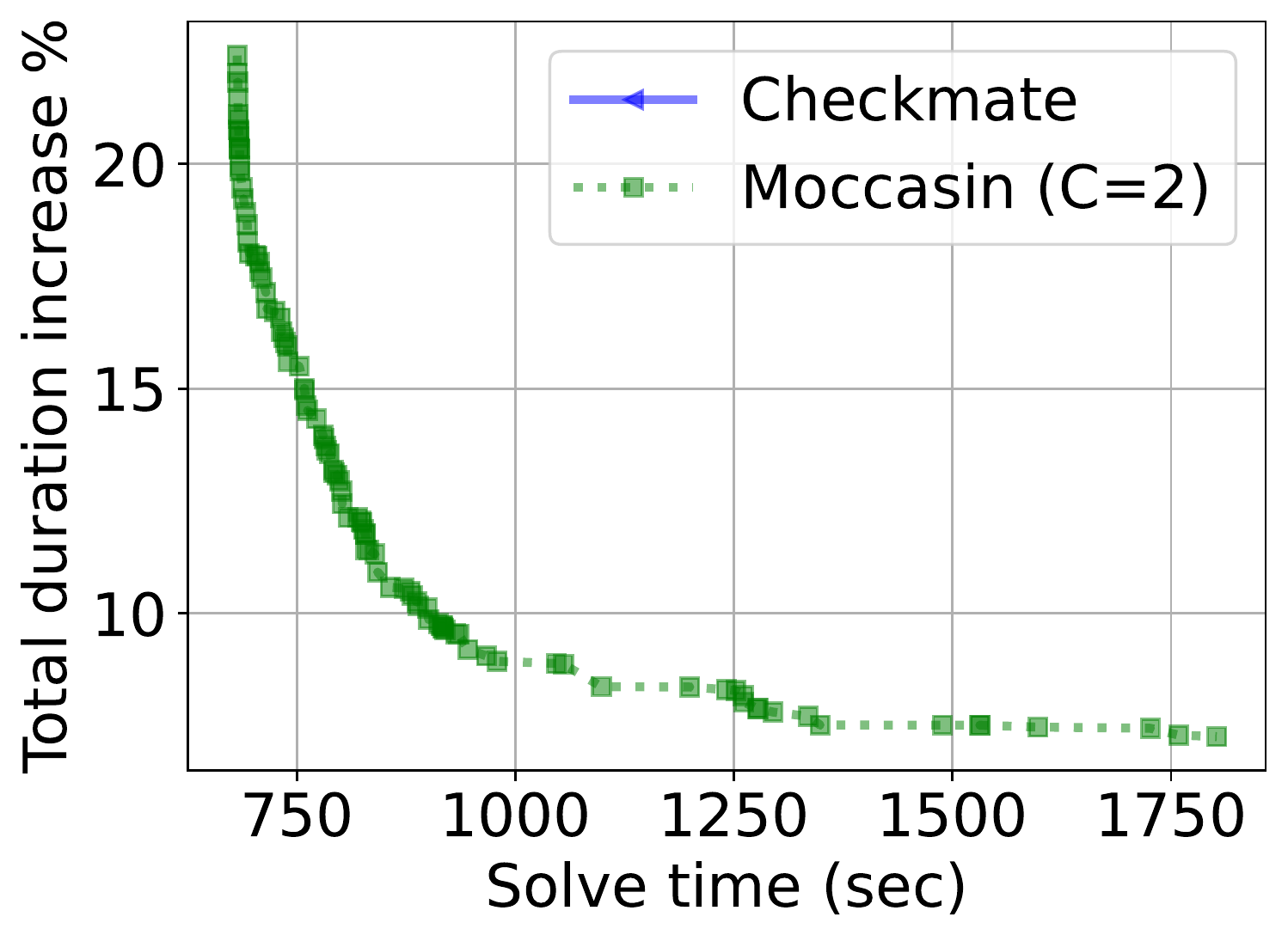}
        \centerline{\scriptsize $G_3, M=213329$}\medskip
    \end{minipage}
    \begin{minipage}{.24\linewidth}
        \centering
        \includegraphics[width=\columnwidth]{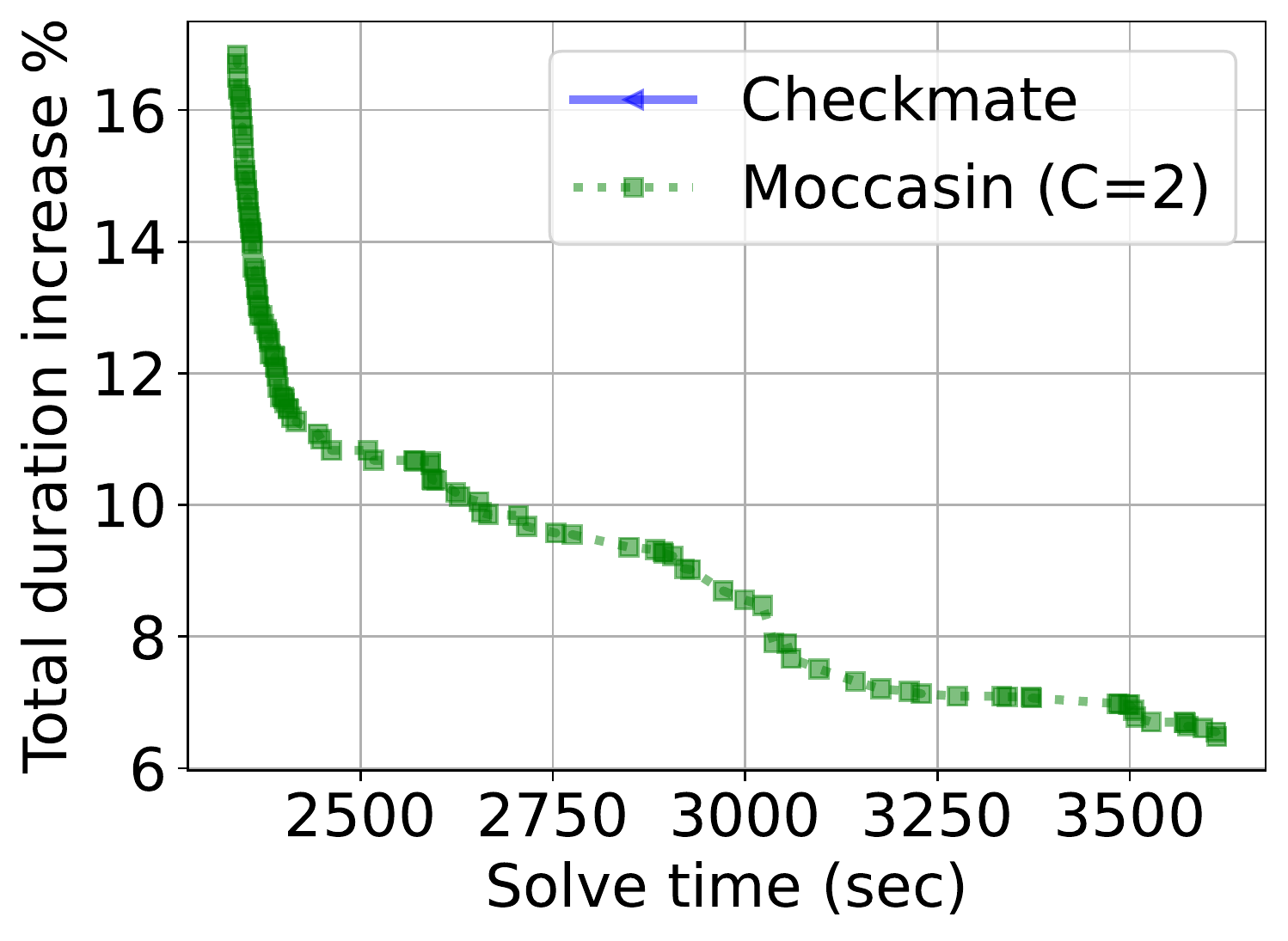}
        \centerline{\scriptsize $G_4, M=456464$}\medskip
    \end{minipage}
    \caption{Solve progress plots for random layered graphs under different memory budget constraints. The graph name and memory budget are as indicated in the caption for each plot. The node and edge counts $(n,m)$ for the graphs are as follows: $G_1: (100, 236), G_2: (250, 944), G_3: (500, 2461), G_4: (1000, 5875)$.}
    \label{fig:solve_progress_randomlayered}
\end{figure*}

Figure \ref{fig:solve_progress_randomlayered} shows the solve progress plots for 4 random layered graphs $G_1,\dots,G_4$ of varied sizes and 4 different local memory budget values for each graph. The curves of \textsc{Moccasin} are all shifted to the right by the amount of time spent in Phase 1 for fair comparison. We have set the time limit to $30$ minutes for $G_1,G_2,G_3$ and $1$ hour for $G_4$.

For $G_1$, the solve times are in the order of a few seconds and we observe that \textsc{Moccasin} is faster to obtain similar objective values as \textsc{Checkmate}. For $G_2$, which has $n=250$ nodes, when the memory budget is tight, we see that \textsc{Checkmate} fails to find a feasible solution within the given time limit of $30$ minutes. For the highest memory budget value, it finds the optimal solution in 10 minutes while \textsc{Moccasin} finishes in a few seconds. For graphs $G_3$ and $G_4$, which have $n=500$ and $n=1000$ nodes respectively, \textsc{Checkmate} times out with no solution even if we consider a higher time limit of 3 hours while \textsc{Moccasin} converges to a good solution (i.e. low total duration increase) under less than an hour. In addition to the solve time, the memory required for optimization also becomes a critical factor for large graphs. In fact, for $G_3$ and $G_4$, \textsc{Checkmate} exits with an out-of-memory error. The results in Figure \ref{fig:solve_progress_randomlayered} are best understood by contrasting the number of discrete variables in \textsc{Checkmate}, which grows quadratically in $n$ while it grows $n\log(n)$ in \textsc{Moccasin} (see Table \ref{table:overview_formulations}).

In all of the experiments, we have set $C_v=2$ for all $v \in V$, which we specified in the plot legends by $C=2$. We have used the version of the formulation in Section \ref{sec:enforce_topo} where we enforce an input topological ordering, which we have randomly generated.

\begin{table*}[htb]
\centering
\caption{Experiment results. The abbreviations used in the table are as follows: RL: Random layered graphs, RW: Real-world graphs, CM: graphs used in \cite{jain2020checkmate} to evaluate \textsc{Checkmate}, where CM 1 is FCN with VGG layers and CM 2 is the ResNet50 model, $n$: number of nodes, $m$: number of edges, $M$: memory budget, TDI: total duration increase in percentage, peak mem: peak memory of the resulting rematerialization sequence. The column `Time (s)' indicates the elapsed time in seconds until the best solution. Dashes ``-" indicate that no solution is found. The best solution in each row is shown in bold font. Full version of this table could be found in the Appendix.}
\vspace{3mm}
\label{table:num_results}
\scalebox{0.83}{
\begin{tabular}{lll|rrr|rrr|rrr} 
 \multicolumn{2}{c}{} & & \multicolumn{3}{c}{\textsc{Checkmate} MILP} & \multicolumn{3}{c}{\textsc{Checkmate} LP+Rounding} & \multicolumn{3}{c}{\textsc{Moccasin}} \\ 
 Graph & $(n,m)$ & $M$ & TDI \% & Peak mem & Time (s) & TDI \% & Peak mem & Time (s) & TDI \% & Peak mem & Time (s) \\ 
 \midrule\midrule
RL $G_2$ & (250, 944) & 132,156 & 0.9 & 132,130 & 685.1 & 93.0 & 178,200 & 401.7 & 0.9 & 131,831 & \textbf{55.0} \\
 &  & 117,472 & - & - & - & 328.6 & 181,200 & 696.9 & 4.9 & 117,400 & \textbf{639.5} \\
RL $G_4$ & (1000, 5875) & 547,757 & - & - & - & - & - & - & 0.7 & 547,660 & \textbf{3612.9} \\
 &  & 486,895 & - & - & - & - & - & - & 3.4 & 486,880 & \textbf{3611.8} \\
\midrule 
RW 1 & (358, 947) & 20,227,276 & 2.3 & 20,226,048 & 1340.0 & - & - & - & 2.3 & 20,226,048 & \textbf{123.5} \\
 &  & 17,979,801 & 4.5 & 17,977,344 & 1605.4 & - & - & - & 4.5 & 17,977,344 & \textbf{122.0} \\
RW 2 & (442, 1247) & 10,817,740 & 1.4 & 10,811,392 & 1856.7 & - & - & - & 1.4 & 10,813,440 & \textbf{1201.3} \\
 &  & 9,615,769 & 2.8 & 9,615,360 & 2242.9 & - & - & - & 2.8 & 9,613,312 & \textbf{303.9} \\
RW 3 & (574, 1304) & 10,539,417 & - & - & - & - & - & - & 0.8 & 10,539,008 & \textbf{1802.4} \\
 &  & 9,368,371 & - & - & - & - & - & - & 1.6 & 9,367,552 & \textbf{1802.8} \\
\midrule 
CM 1 & (73, 149) & 11.3 GB & 0.0 & 11.1 GB & 6.3 & 0.0 & 11.4 GB & 6.8 & 0.0 & 11.1 GB & \textbf{3.1} \\
 &  & 10.0 GB & 0.1 & 9.65 GB & 5.6 & 0.1 & 10.8 GB & 6.7 & 0.1 & 9.9 GB & \textbf{3.1} \\
CM 2 & (353, 751) & 31.9 GB & \textbf{0.1} & 31.6 GB & 434.1 & 0.1 & 31.5 GB & 505.2 & 0.2 & 31.9 GB & \textbf{65.2} \\
 &  & 28.4 GB & 0.3 & 28.3 GB & 485.3 & 0.5 & 27.8 GB & 1065.4 & 0.3 & 28.4 GB & \textbf{69.3} \\
\end{tabular}
}
\end{table*}

Table \ref{table:num_results} provides numerical results for a range of different computation graphs. In Table \ref{table:num_results} we have selected the memory budget values for each graph to be the $80\%$ and $90\%$ of the initial peak memory without rematerialization. In addition to the methods of \textsc{Checkmate} and \textsc{Moccasin}, we include results for the rounding algorithm proposed in \cite{jain2020checkmate} under the ``LP+rounding" column of the table. This method consists of relaxing the MILP into a linear program (LP) and then rounding the solution (see \cite{jain2020checkmate} for further details on this algorithm). Note that solution produced by the rounding algorithm is not guaranteed to satisfy the memory budget constraint. This could be seen in Table \ref{table:num_results} where in most cases the peak memory for the relaxation and rounding approach is higher than the memory budget $M$.

Table \ref{table:num_results} shows that the random layered graphs and real-world graphs are the most challenging ones among the graph set. The solve times for these graphs are higher than the CM graphs, which is consistent with the fact that they have higher edge densities and more complex edge connectivities. 

It is critical to highlight that \textsc{Moccasin} is able to scale to graphs with few hundreds of nodes and a few thousands of edges. This is while achieving a few percents of total duration increase, which is essential to performance.

\section{Conclusion} \label{sec:conclusion}

Rematerialization is a valuable technique that allows to save peak local memory footprint at the expense of longer total duration of computation. However, finding a rematerialization sequence that minimizes total duration while meeting a prescribed local memory limit is not easy - especially for graphs with complex interconnections.  Previous research \cite{jain2020checkmate} demonstrates a MILP formulation that can address this optimization problem, however our experiments show that that approach does not scale well as the graph size, and the complexity of graph topology grow.

We introduce \textsc{Moccasin}, a new formulation that expresses the decision variables as \emph{retention intervals} specified by the start and end event indices of each node's output tensor.  We further manage complexity using the hyper-parameter $C_v$ which limits the maximum number of retention intervals for each node.  Our experimental results include \textsc{Checkmate} graphs (training graphs with simple interconnect topology), synthetically generated random layered graphs (that model inference graphs with complex topology), and real-world inference graphs in active use commercially.  We demonstrate that \textsc{Moccasin} provides the same optimization result as \textsc{Checkmate} with up to an order-of-magnitude less solve time for mid-sized graphs (100-250 nodes).  It also enables us to find solutions for larger graphs (up to 1000 nodes) and for more challenging peak local memory limits, cases where the MILP formulation of \textsc{Checkmate} failed to return any solution within the time limit.

Finally, rematerialization is seen as one component among a set of tools for managing memory use during execution of complex computation graphs arising in deep learning applications.  Joint optimization of this with other methods such as sequencing, scheduling for parallel compute, and paging to global memory are viewed as valuable topics for further research.



\bibliography{biblio}

\begin{thebibliography}{28}
\providecommand{\natexlab}[1]{#1}
\providecommand{\url}[1]{\texttt{#1}}
\expandafter\ifx\csname urlstyle\endcsname\relax
  \providecommand{\doi}[1]{doi: #1}\else
  \providecommand{\doi}{doi: \begingroup \urlstyle{rm}\Url}\fi

\bibitem[Agrawal et~al.(2018)Agrawal, Verschueren, Diamond, and
  Boyd]{agrawal2018rewriting}
Agrawal, A., Verschueren, R., Diamond, S., and Boyd, S.
\newblock A rewriting system for convex optimization problems.
\newblock \emph{Journal of Control and Decision}, 5\penalty0 (1):\penalty0
  42--60, 2018.

\bibitem[Beaumont et~al.(2021)Beaumont, Eyraud-Dubois, and
  Shilova]{beaumont2021efficient}
Beaumont, O., Eyraud-Dubois, L., and Shilova, A.
\newblock Efficient combination of rematerialization and offloading for
  training dnns.
\newblock \emph{Advances in Neural Information Processing Systems},
  34:\penalty0 23844--23857, 2021.

\bibitem[Briggs et~al.(1992)Briggs, Cooper, and
  Torczon]{briggs1992rematerialization}
Briggs, P., Cooper, K.~D., and Torczon, L.
\newblock Rematerialization.
\newblock In \emph{Proceedings of the ACM SIGPLAN 1992 conference on
  Programming language design and implementation}, pp.\  311--321, 1992.

\bibitem[Chen et~al.(2016)Chen, Xu, Zhang, and Guestrin]{chen2016training}
Chen, T., Xu, B., Zhang, C., and Guestrin, C.
\newblock Training deep nets with sublinear memory cost.
\newblock \emph{arXiv preprint arXiv:1604.06174}, 2016.

\bibitem[Colombet et~al.(2015)Colombet, Brandner, and
  Darte]{colombet2015studying}
Colombet, Q., Brandner, F., and Darte, A.
\newblock Studying optimal spilling in the light of ssa.
\newblock \emph{ACM Transactions on Architecture and Code Optimization (TACO)},
  11\penalty0 (4):\penalty0 1--26, 2015.

\bibitem[Diamond \& Boyd(2016)Diamond and Boyd]{diamond2016cvxpy}
Diamond, S. and Boyd, S.
\newblock {CVXPY}: {A} {P}ython-embedded modeling language for convex
  optimization.
\newblock \emph{Journal of Machine Learning Research}, 17\penalty0
  (83):\penalty0 1--5, 2016.

\bibitem[Gagrani et~al.(2022)Gagrani, Rainone, Yang, Teague, Jeon, Bondesan,
  van Hoof, Lott, Zeng, and Zappi]{topoformer}
Gagrani, M., Rainone, C., Yang, Y., Teague, H., Jeon, W., Bondesan, R., van
  Hoof, H., Lott, C., Zeng, W.~W., and Zappi, P.
\newblock Neural topological ordering for computation graphs.
\newblock In \emph{Advances in Neural Information Processing Systems}, 2022.

\bibitem[Gilbert et~al.(1979)Gilbert, Lengauer, and
  Tarjan]{gilbert1979pebbling}
Gilbert, J.~R., Lengauer, T., and Tarjan, R.~E.
\newblock The pebbling problem is complete in polynomial space.
\newblock In \emph{Proceedings of the eleventh annual ACM symposium on Theory
  of computing}, pp.\  237--248, 1979.

\bibitem[Griewank \& Walther(2000)Griewank and Walther]{griewank2000algorithm}
Griewank, A. and Walther, A.
\newblock Algorithm 799: revolve: an implementation of checkpointing for the
  reverse or adjoint mode of computational differentiation.
\newblock \emph{ACM Transactions on Mathematical Software (TOMS)}, 26\penalty0
  (1):\penalty0 19--45, 2000.

\bibitem[Griewank \& Walther(2008)Griewank and Walther]{griewank2008evaluating}
Griewank, A. and Walther, A.
\newblock \emph{Evaluating derivatives: principles and techniques of
  algorithmic differentiation}.
\newblock SIAM, 2008.

\bibitem[{Gurobi Optimization, LLC}(2023)]{gurobi}
{Gurobi Optimization, LLC}.
\newblock {Gurobi Optimizer Reference Manual}, 2023.
\newblock URL \url{https://www.gurobi.com}.

\bibitem[Huang et~al.(2019)Huang, Cheng, Bapna, Firat, Chen, Chen, Lee, Ngiam,
  Le, Wu, et~al.]{huang2019gpipe}
Huang, Y., Cheng, Y., Bapna, A., Firat, O., Chen, D., Chen, M., Lee, H., Ngiam,
  J., Le, Q.~V., Wu, Y., et~al.
\newblock Gpipe: Efficient training of giant neural networks using pipeline
  parallelism.
\newblock \emph{Advances in neural information processing systems}, 32, 2019.

\bibitem[Jain(2020)]{checkmate_repo}
Jain, P.
\newblock Checkmate repository, 2020.
\newblock URL \url{https://github.com/parasj/checkmate/tree/mlsys20_artifact}.

\bibitem[Jain et~al.(2020)Jain, Jain, Nrusimha, Gholami, Abbeel, Gonzalez,
  Keutzer, and Stoica]{jain2020checkmate}
Jain, P., Jain, A., Nrusimha, A., Gholami, A., Abbeel, P., Gonzalez, J.,
  Keutzer, K., and Stoica, I.
\newblock Checkmate: Breaking the memory wall with optimal tensor
  rematerialization.
\newblock \emph{Proceedings of Machine Learning and Systems}, 2:\penalty0
  497--511, 2020.

\bibitem[Kirisame et~al.(2021)Kirisame, Lyubomirsky, Haan, Brennan, He, Roesch,
  Chen, and Tatlock]{kirisame2021dynamic}
Kirisame, M., Lyubomirsky, S., Haan, A., Brennan, J., He, M., Roesch, J., Chen,
  T., and Tatlock, Z.
\newblock Dynamic tensor rematerialization.
\newblock In \emph{International Conference on Learning Representations}, 2021.
\newblock URL \url{https://openreview.net/forum?id=Vfs_2RnOD0H}.

\bibitem[Kubota(1998)]{kubota1998fortran77}
Kubota, K.
\newblock A fortran77 preprocessor for reverse mode automatic differentiation
  with recursive checkpointing.
\newblock \emph{Optimization Methods and Software}, 10\penalty0 (2):\penalty0
  319--335, 1998.

\bibitem[Kumar et~al.(2019)Kumar, Purohit, Svitkina, Vee, and
  Wang]{kumar2019efficient}
Kumar, R., Purohit, M., Svitkina, Z., Vee, E., and Wang, J.
\newblock Efficient rematerialization for deep networks.
\newblock \emph{Advances in Neural Information Processing Systems}, 32, 2019.

\bibitem[Kusumoto et~al.(2019)Kusumoto, Inoue, Watanabe, Akiba, and
  Koyama]{kusumoto2019graph}
Kusumoto, M., Inoue, T., Watanabe, G., Akiba, T., and Koyama, M.
\newblock A graph theoretic framework of recomputation algorithms for
  memory-efficient backpropagation.
\newblock \emph{Advances in Neural Information Processing Systems}, 32, 2019.

\bibitem[Lozano \& Schulte(2019)Lozano and Schulte]{lozano2019survey}
Lozano, R.~C. and Schulte, C.
\newblock Survey on combinatorial register allocation and instruction
  scheduling.
\newblock \emph{ACM Computing Surveys (CSUR)}, 52\penalty0 (3):\penalty0 1--50,
  2019.

\bibitem[Lozano et~al.(2019)Lozano, Carlsson, Blindell, and
  Schulte]{lozano2019combinatorial}
Lozano, R.~C., Carlsson, M., Blindell, G.~H., and Schulte, C.
\newblock Combinatorial register allocation and instruction scheduling.
\newblock \emph{ACM Transactions on Programming Languages and Systems
  (TOPLAS)}, 41\penalty0 (3):\penalty0 1--53, 2019.

\bibitem[Mostafa(2022)]{mostafa2022sequential}
Mostafa, H.
\newblock Sequential aggregation and rematerialization: Distributed full-batch
  training of graph neural networks on large graphs.
\newblock \emph{Proceedings of Machine Learning and Systems}, 4:\penalty0
  265--275, 2022.

\bibitem[Ohrimenko et~al.(2009)Ohrimenko, Stuckey, and
  Codish]{ohrimenko2009propagation}
Ohrimenko, O., Stuckey, P.~J., and Codish, M.
\newblock Propagation via lazy clause generation.
\newblock \emph{Constraints}, 14\penalty0 (3):\penalty0 357--391, 2009.

\bibitem[Patil et~al.(2022)Patil, Jain, Dutta, Stoica, and
  Gonzalez]{patil2022poet}
Patil, S.~G., Jain, P., Dutta, P., Stoica, I., and Gonzalez, J.
\newblock Poet: Training neural networks on tiny devices with integrated
  rematerialization and paging.
\newblock In \emph{International Conference on Machine Learning}, pp.\
  17573--17583. PMLR, 2022.

\bibitem[Perron \& Furnon(2022)Perron and Furnon]{ortools}
Perron, L. and Furnon, V.
\newblock Or-tools, 2022.
\newblock URL \url{https://developers.google.com/optimization/}.

\bibitem[Schuler et~al.(2022)Schuler, Membarth, and
  Slusallek]{schuler2022xengine}
Schuler, M., Membarth, R., and Slusallek, P.
\newblock Xengine: Optimal tensor rematerialization for neural networks in
  heterogeneous environments.
\newblock \emph{ACM Transactions on Architecture and Code Optimization}, pp.\
  ~5, 2022.

\bibitem[Schutt et~al.(2013)Schutt, Feydy, Stuckey, and
  Wallace]{schutt2013solving}
Schutt, A., Feydy, T., Stuckey, P.~J., and Wallace, M.~G.
\newblock Solving rcpsp/max by lazy clause generation.
\newblock \emph{Journal of scheduling}, 16\penalty0 (3):\penalty0 273--289,
  2013.

\bibitem[Sze et~al.(2017)Sze, Chen, Yang, and Emer]{sze2017efficient}
Sze, V., Chen, Y.-H., Yang, T.-J., and Emer, J.~S.
\newblock Efficient processing of deep neural networks: A tutorial and survey.
\newblock \emph{Proceedings of the IEEE}, 105\penalty0 (12):\penalty0
  2295--2329, 2017.

\bibitem[tinyML Foundation(2022)]{tinyML_OnDevice_Form}
tinyML Foundation.
\newblock tiny{ML} {O}n-{D}evice {L}earning {F}orum, 2022.
\newblock URL \url{https://www.tinyml.org/event/on-device-learning/}.

\end{thebibliography}
\bibliographystyle{icml2023}

\newpage
\appendix
\onecolumn


\section{Appendix} \label{sec:appendix}
\subsection{Additional Numerical Experiments}
\begin{figure}[h]
    \centering
    \includegraphics[width=0.35\columnwidth]{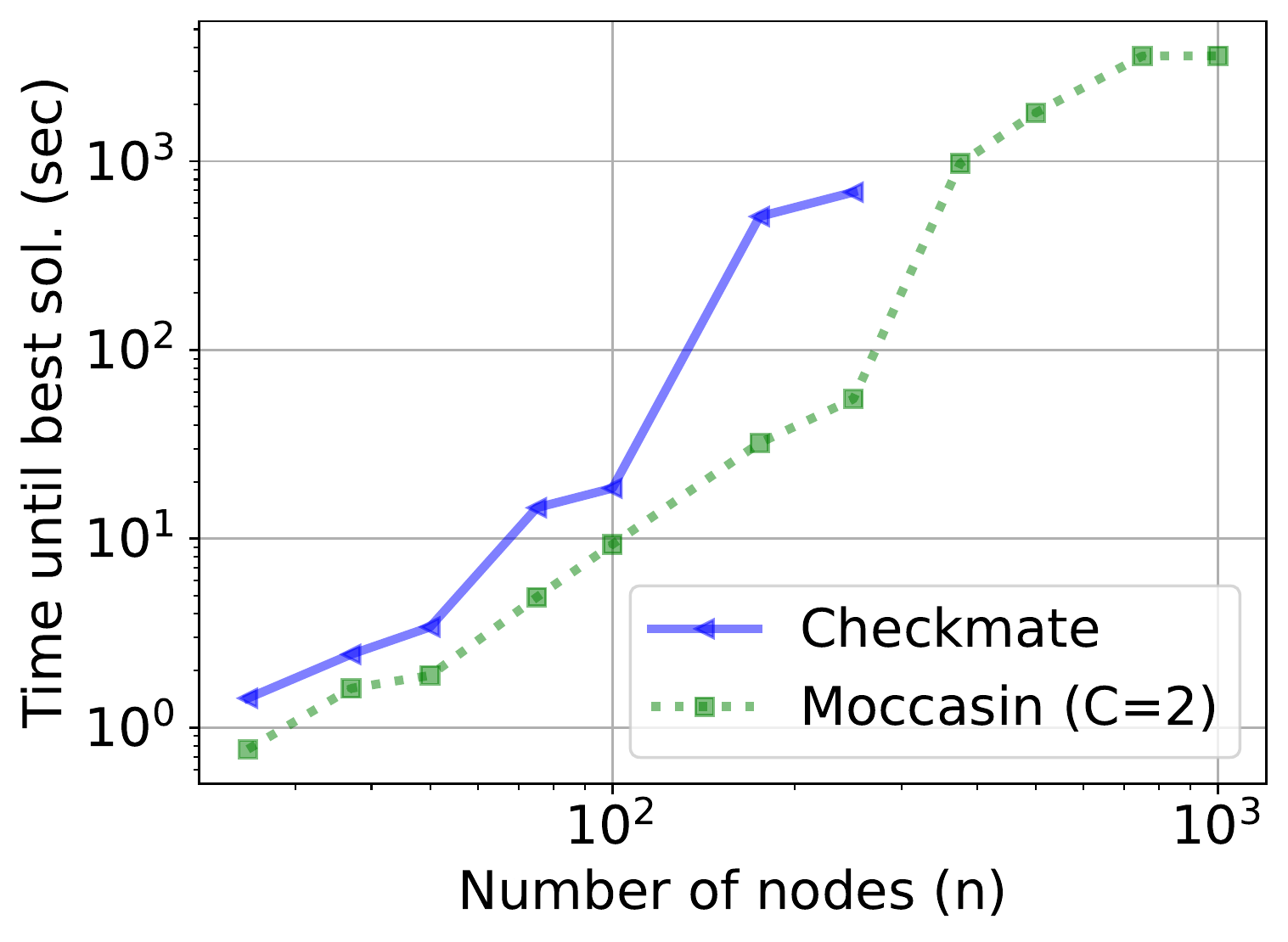}
    \caption{Time until the best solution as a function of number of nodes $n$ for random layered graphs. The memory budget is selected for each graph as the $90\%$ of the peak memory for the input topological order.}
    \label{fig:scalability_plot}
\end{figure}

Figure \ref{fig:scalability_plot} shows the time until best solution as a function of number of nodes $n$ in log-log scale. The reason why the blue curve stops after $n=250$ is that \textsc{Checkmate} does not produce a feasible solution within the given time limit of 1 hour. Figure \ref{fig:scalability_plot} shows that our method demonstrates better scalability. In instances where both methods return feasible solutions, the objective function values are the same. For graphs with more than $n=250$ nodes, for the feasible solutions that our method finds, the corresponding total duration increase is consistently less than $5\%$.

Table \ref{table:num_results_full} shows the extended version of Table \ref{table:num_results}, where results for additional graphs are included.

\begin{table*}[h]
\centering
\caption{Extended version of Table \ref{table:num_results}.}
\vspace{3mm}
\label{table:num_results_full}
\scalebox{0.83}{
\begin{tabular}{lll|rrr|rrr|rrr} 
 \multicolumn{2}{c}{} & & \multicolumn{3}{c}{\textsc{Checkmate} MILP} & \multicolumn{3}{c}{\textsc{Checkmate} LP+Rounding} & \multicolumn{3}{c}{\textsc{Moccasin}} \\ 
 Graph & $(n,m)$ & $M$ & TDI \% & Peak mem & Time (s) & TDI \% & Peak mem & Time (s) & TDI \% & Peak mem & Time (s) \\ 
 \midrule\midrule
RL $G_1$ & (100, 236) & 41,687 & 0.8 & 41,360 & 18.5 & 0.1 & 44,939 & 16.0 & 0.8 & 40,820 & 9.3 \\
 &  & 37,055 & 2.3 & 36,830 & 22.7 & 7.3 & 43,370 & 20.1 & 2.3 & 36,810 & 9.5 \\
RL $G_2$ & (250, 944) & 132,156 & 0.9 & 132,130 & 685.1 & 93.0 & 178,200 & 401.7 & 0.9 & 131,831 & 55.0 \\
 &  & 117,472 & - & - & - & 328.6 & 181,200 & 696.9 & 4.9 & 117,400 & 639.5 \\
RL $G_3$ & (500, 2461) & 255,995 & - & - & - & - & - & - & 0.7 & 255,959 & 1803.3 \\
 &  & 227,551 & - & - & - & - & - & - & 3.4 & 227,429 & 1804.8 \\
RL $G_4$ & (1000, 5875) & 547,757 & - & - & - & - & - & - & 0.7 & 547,660 & 3612.9 \\
 &  & 486,895 & - & - & - & - & - & - & 3.4 & 486,880 & 3611.8 \\
\midrule 
RW 1 & (358, 947) & 20,227,276 & 2.3 & 20,226,048 & 1340.0 & - & - & - & 2.3 & 20,226,048 & 123.5 \\
 &  & 17,979,801 & 4.5 & 17,977,344 & 1605.4 & - & - & - & 4.5 & 17,977,344 & 122.0 \\
RW 2 & (442, 1247) & 10,817,740 & 1.4 & 10,811,392 & 1856.7 & - & - & - & 1.4 & 10,813,440 & 1201.3 \\
 &  & 9,615,769 & 2.8 & 9,615,360 & 2242.9 & - & - & - & 2.8 & 9,613,312 & 303.9 \\
RW 3 & (574, 1304) & 10,539,417 & - & - & - & - & - & - & 0.8 & 10,539,008 & 1802.4 \\
 &  & 9,368,371 & - & - & - & - & - & - & 1.6 & 9,367,552 & 1802.8 \\
RW 4 & (698, 1436) & 24,328,396 & - & - & - & - & - & - & 1.7 & 24,328,192 & 1803.9 \\
 &  & 21,625,241 & - & - & - & - & - & - & 3.4 & 21,624,832 & 1314.0 \\
\midrule 
CM 1 & (73, 149) & 11.3 GB & 0.0 & 11.1 GB & 6.3 & 0.0 & 11.4 GB & 6.8 & 0.0 & 11.1 GB & 3.1 \\
 &  & 10.0 GB & 0.1 & 9.65 GB & 5.6 & 0.1 & 10.8 GB & 6.7 & 0.1 & 9.9 GB & 3.1 \\
CM 2 & (353, 751) & 31.9 GB & 0.1 & 31.6 GB & 434.1 & 0.1 & 31.5 GB & 505.2 & 0.2 & 31.9 GB & 65.2 \\
 &  & 28.4 GB & 0.3 & 28.3 GB & 485.3 & 0.5 & 27.8 GB & 1065.4 & 0.3 & 28.4 GB & 69.3 \\
\end{tabular}
}
\end{table*}

\subsection{Examples of Graphs Used in Experiments} \label{sec:app_graphs}
\begin{figure*}[hbt!]
    \centering
        
    \begin{minipage}{.40\linewidth}
        \centering
        \includegraphics[width=\columnwidth]{src/Checkmate_fcn_8_vgg_MLSys_32_416_608_3_train_nx.simple.pdf}
         \centerline{\scriptsize Checkmate FCN8 (73 nodes, 149 edges)}\medskip
    \end{minipage}%
    \begin{minipage}{.40\linewidth}
        \centering
        \includegraphics[width=\columnwidth]{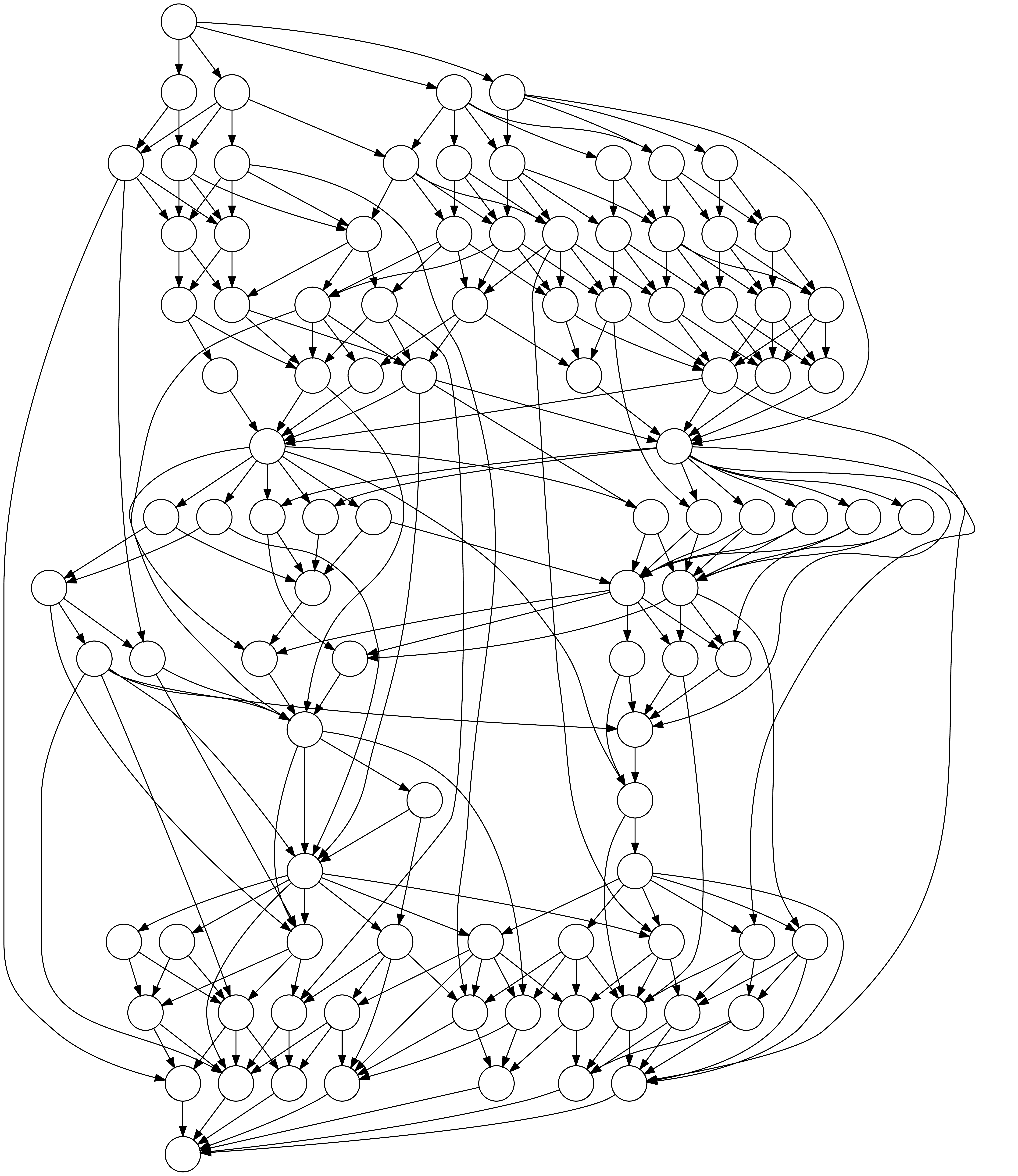}
         \centerline{\scriptsize Random layered example (100 nodes, 268 edges)}\medskip
    \end{minipage}

    \vspace{-3mm}
    \caption{The example graphs used for our experiments are shown using a simple visualization showing only the structure of the graphs.}
    \label{fig:example_graphs}
\end{figure*}

Figure \ref{fig:example_graphs} shows visualizations of two of the graphs that we use for our experiments to give a better understanding of the structure of these graphs.  The Checkmate FCN8 graph (left) is a training computation graph and the 100-node random layered graph (right) models an inference graph. We study much larger graphs, but these are quite difficult to visualize in two dimensions.  

\end{document}